\documentclass[]{fairmeta}
\usepackage{amssymb}    
\usepackage{pifont}     
\usepackage[table]{xcolor}  
\usepackage{booktabs}   %
\usepackage[inline]{enumitem} %
\usepackage{epigraph}
\usepackage{subcaption}   

\usepackage{amsmath,amsfonts,bm}

\def\eqref#1{equation~\ref{#1}}

\def\1{\bm{1}}

\def\mA{{\bm{A}}}

\def\mI{{\bm{I}}}

\def\mT{{\bm{T}}}

\DeclareMathAlphabet{\mathsfit}{\encodingdefault}{\sfdefault}{m}{sl}
\SetMathAlphabet{\mathsfit}{bold}{\encodingdefault}{\sfdefault}{bx}{n}
\newcommand{\tens}[1]{\bm{\mathsfit{#1}}}

\def\tG{{\tens{G}}}

\def\tV{{\tens{V}}}

\def\tX{{\tens{X}}}

\def\tZ{{\tens{Z}}}

\usepackage{xcolor}
\usepackage{wrapfig}
\usepackage{booktabs} 
\usepackage{tabularx}
\usepackage{titletoc}
\definecolor{skyblue}{RGB}{0, 162, 232}   
\definecolor{actionred}{RGB}{220, 30, 30}  

\title{World Action Modeling with Progressive Visual Planning
}

\author[1,2,3*]{Fei Zhang}
\author[3]{Zhaochong An}
\author[3]{Duncan Frost}
\author[3]{Yikai Wang}
\author[1,2]{Pengfei Liu}
\author[1]{Ya Zhang}
\author[3]{Michal Drozdzal}
\author[4]{Amir Bar}

\affiliation[1]{Shanghai Jiao Tong University}
\affiliation[2]{SII}
\affiliation[3]{Meta}
\affiliation[4]{Imperial College London}

\contribution[*]{Work done at Meta}

\abstract{\emph{World action models} (WAMs) have emerged as a promising paradigm for robotic control by jointly predicting future visual dynamics and actions from an initial observation and instruction. However, existing WAMs struggle with long-horizon prediction, as generating dense video rollouts is highly inefficient. Some recent WAMs address this by predicting a single future frame without generating the full video, but this approach neglects \emph{how to progress} toward the goal. We present \textbf{ProWAM}, a progressive world action model that jointly predicts actions and an ordered sequence of sparse visual sub-goals, providing explicit visual guidance to anchor action generation throughout task execution. This design scales naturally, as sub-goal prediction can be learned from large-scale action-free videos, allowing the video backbone to offload complex visual planning from the action policy. For efficient action generation, ProWAM executes a single video-backbone forward pass to cache sparse sub-goal features, eliminating iterative full-video generation and requiring only lightweight action denoising during replanning. Across extensive evaluations, ProWAM achieves superior out-of-distribution robustness. On simulation benchmarks, it sets new \emph{state-of-the-art} results on LIBERO-Plus (85.8\%) and randomized RoboTwin (75.7\%), outperforming the strongest baseline with relative gains of up to \textbf{+35.9\%}. On RoboCasa365, ProWAM achieves a 48.1\% success rate and 18.2\% on the challenging Composite-Unseen split, ranking \textbf{4th overall}. Crucially, in zero-shot real-world experiments, ProWAM achieves 70.0\% success, outperforming the strongest baseline by \textbf{+15.0} (from 55.0\% to 70.0\%, a \textbf{+27.3\%} relative gain) in novel scenes. These results demonstrate the value of progress-indexed visual foresight for closed-loop control.
}

\date{\today}

\metadata[Project Page]{\url{https://sii-ferenas.github.io/ProWAM-page}}

\begin{document}

\maketitle

\section{Introduction}\label{sec_intro}

\begin{figure*}[t]
  \centering
  \vspace{-5mm}
  \includegraphics[width=1.0\textwidth]{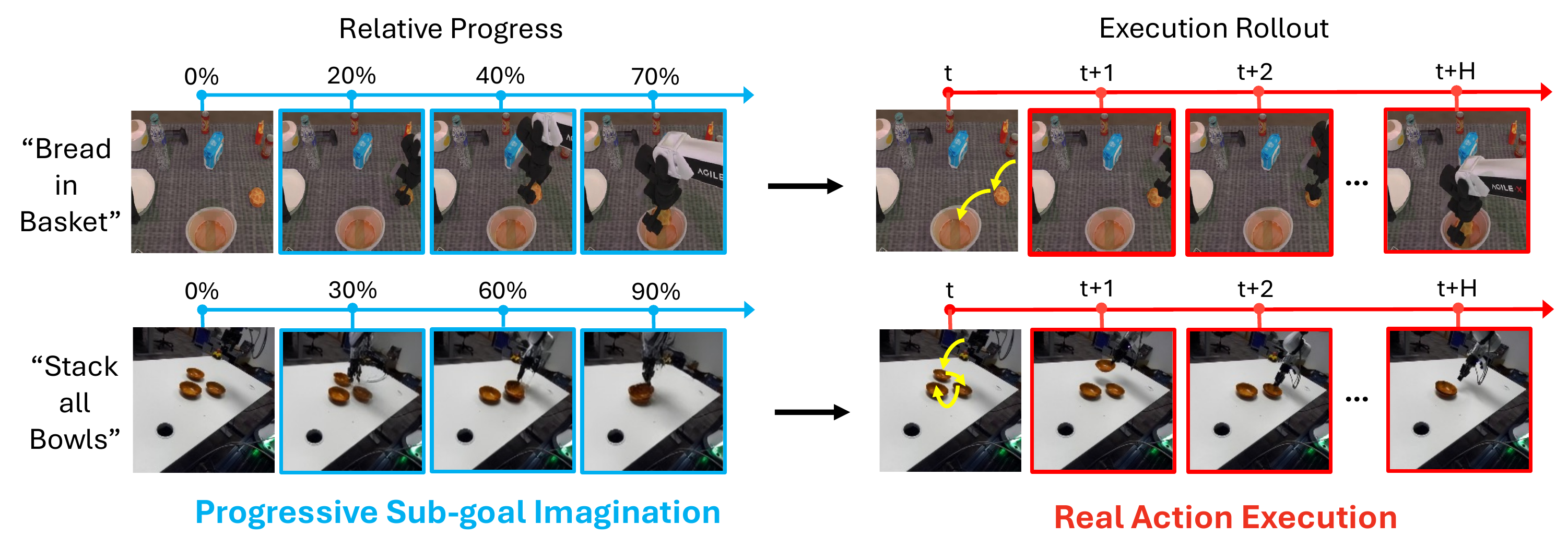}
  \vspace{-7mm}
  \caption{\textbf{Progressive Sub-goal
  \textcolor{skyblue}{Imagination} for Long-Horizon
  \textcolor{actionred}{Control}.}
  Given the current observation, ProWAM generates sparse visual targets at
  requested relative progress values (\emph{left}). The resulting sub-goal
  features directly condition action generation for the corresponding execution
  rollout (\emph{right}), providing explicit visual guidance without
  generating dense video rollouts.}
\label{fig:teaser}
\end{figure*}

\emph{Vision-language-action} (VLA) models primarily inherit semantic knowledge from vision-language pretraining but remain data-hungry due to their reliance on large, costly collections of robot demonstrations~\citep{kim2024openvla,24pi0,25pi0.5}. In contrast, \emph{world action models} (WAMs) introduce a complementary source of transferable knowledge: visual-dynamics priors learned by video generation models from large-scale video datasets~\citep{blattmann2023svd,yang2024cogvideox,wan2025wan}. Recent WAMs have shown that incorporating such predictive visual knowledge can improve data efficiency and out-of-distribution generalization in robotic control~\citep{pai2025mimic,ye2026dreamzero,yuan2026fastwam,zhang2026imagewam,li2026lingbotva}.

Existing WAMs differ in how they utilize future visual prediction to guide action generation. \emph{Full-imagination} WAMs~\citep{pai2025mimic,ye2026dreamzero,li2026lingbotva} leverage dense future rollouts that provide detailed near-term guidance but are inefficient as they require multiple denoising passes through a large video backbone. Alternatively, \emph{zero-imagination} WAMs~\citep{yuan2026fastwam,ye2026gigaworldpolicy} generate actions without predicting future visual states. Between these extremes, ImageWAM~\citep{zhang2026imagewam} conditions control on an action-horizon endpoint frame without explicitly representing intermediate visual states. These designs reveal a fundamental trade-off: dense visual rollouts provide short-horizon dynamics at high computational cost, whereas other efficient alternatives lack explicit intermediate-progress guidance over the full task.

To reconcile this trade-off, we introduce \textbf{ProWAM}, a \emph{progressive world action model} that extends a single fixed-state prediction into an ordered sequence of sparse visual sub-goals. Each sub-goal is indexed by normalized progress relative to the current state, where $r=0$ denotes the current observation and $r=1$ denotes the trajectory endpoint. This relative parameterization dynamically re-anchors sub-goals to the current observation, enabling consistent guidance across arbitrary task stages and varying trajectory lengths. Consequently, these sparse targets efficiently guide action generation without requiring dense video rollouts or separate high-level planning modules.

Specifically, ProWAM adopts a vision-action \emph{mixture-of-transformers} (MoT) architecture consisting of a video generation expert and a lightweight diffusion-based action expert for policy generation. Each visual sub-goal is represented as a latent slot associated with a normalized progress value. The video expert jointly predicts near-term future transitions and progress-conditioned sub-goals, while the action expert  attends to current observations and sub-goals to generate low-level action chunks. Asymmetric attention prevents action information from altering the visual prediction pathway. This design preserves the generative prior of the video expert, enabling video-only pretraining on large-scale action-free videos followed by joint fine-tuning on task-specific robot demonstrations to couple visual planning with action generation. 
At inference, ProWAM reuses cached sub-goal features to eliminate repeated video-backbone computation, enabling efficient closed-loop control. 
Overall, our contributions are threefold:

\begin{itemize}[label=$\bullet$, leftmargin=1.5em, itemsep=2pt, topsep=2pt]

\item We introduce \emph{progressive sub-goal imagination} (Figure~\ref{fig:teaser}), a new visual planning paradigm for WAMs that represents long-horizon futures as an ordered sequence of sparse visual sub-goals indexed by normalized task progress. This representation provides explicit intermediate guidance without requiring dense video rollouts during inference or an external planner.

\item We instantiate this paradigm in \textbf{ProWAM}, a WAM that natively integrates progress-conditioned visual sub-goal prediction with action generation for long-horizon control. By embedding sub-goal slots into the video pathway and coupling them via asymmetric attention, ProWAM enables scalable action-free video pretraining followed by joint vision-action fine-tuning. At inference, cached sub-goal features guide action denoising without repeated video-expert computation.




\item Extensive simulation and real-world evaluations demonstrate that ProWAM achieves exceptional out-of-distribution robustness, setting a new \emph{state-of-the-art} on LIBERO-Plus~\citep{fei25libero-plus} and randomized RoboTwin~\citep{chen2025robotwin} with an average absolute performance gain of \textbf{+11.4} over prior WAMs. Furthermore, it ranks as the \textbf{second} overall on challenging long-horizon RoboCasa365~\citep{nasiriany2024robocasa} benchmark, and outperforms DreamZero~\citep{ye2026dreamzero} by \textbf{+15.0} on zero-shot real-world tasks in novel scenes.
\end{itemize}

\section{Preliminaries}\label{sec_prelim}

\paragraph{\textbf{Diffusion-based Video Generation Pipeline.}} 
Prevailing video generation pipelines~\citep{yang2024cogvideox,wan2025wan} predominantly adopt a continuous-time denoising process formalized via flow matching~\citep{lipman2022flow, esser2024scaling}. In this work, we build upon the I2V framework of Wan~\citep{wan2025wan}. Formally, an input video $\tV$ is first encoded into a compressed latent representation using a pretrained \emph{variational autoencoder} (VAE): $\mathcal{E}(\tV) = \tX \in \mathbb{R}^{c \times (f/4) \times (h/8) \times (w/8)}$, where $c$ is the latent channel dimension, and $f, h, w$ represent the frame count, height, and width of the input video, respectively. Subsequently, we employ flow matching to learn a continuous-time generative diffusion process, which models the mapping between the target latent $\tX^1$ (i.e., $\tX$) and random Gaussian noise $\epsilon \sim \mathcal{N}(0, I)$ via a fixed, linearly sampled timestep process: $\tX^{t} = t\tX^{1} + (1-t)\epsilon, t \in [0,1]$. In this way, a $\theta$-parameterized velocity prediction model $\mathcal{V}: \mathbb{R}^{c \times (f/4) \times (h/8) \times (w/8)} \rightarrow \mathbb{R}^{c \times (f/4) \times (h/8) \times (w/8)}$, instantiated as a \emph{diffusion transformer} (DiT), is trained using the following \emph{mean squared error} (MSE) objective:
\begin{equation}
    \mathcal{L}_{\mathrm{video}} = \mathbb{E}_{\epsilon, \tX^{1}, \mT_\mathrm{ref}, \mI_{\mathrm{ref}}, t}\left\|\mathcal{V} (\tX^{t}, t, \mT_\mathrm{ref}, \mI_{\mathrm{ref}};\theta) - \frac{d\tX^{t}}{dt}\right\|_2^2,
\end{equation}
where the target velocity is given by $d\tX^{t}/dt = \tX^{1} - \epsilon$. Here $\mT_\mathrm{ref}$ and $\mI_\mathrm{ref}$ denote text and image embeddings extracted by off-the-shelf encoders (umT5~\citep{raffel2020exploring} and CLIP~\citep{clip}, respectively). These embeddings are fused with the original latent representation via cross-attention, providing conditional reference for controllable video generation. During inference, a video can be generated by iteratively denoising $\epsilon$ using $\mathcal{V}$, conditioned on the $\mT_\mathrm{ref}$ and $\mI_\mathrm{ref}$.

\paragraph{\textbf{Tailoring Video Generative Knowledge to Physical Action.}}
To repurpose a pretrained video generator as a control policy, a \emph{world action model} (WAM) augments the video backbone with a lightweight action expert, typically instantiated as a DiT~\citep{25pi0.5,ye2026dreamzero}. This expert inherits the visual-dynamics prior of the video model while emitting executable controls. 
Concretely, let
    $\mA=[\mathbf{a}_0,\ldots,\mathbf{a}_{H-1}]
    \in\mathbb{R}^{H \times d_a}$
    denote an action chunk of horizon $H$, where $d_a$ is the action
    dimension. When observations and actions are temporally aligned, a clip
    containing $f$ observation frames has $f-1$ transitions, and we therefore set
    $H=f-1$. The action and video experts jointly learn a policy that generates the action chunk
  conditioned on the textual instruction $\mT_\mathrm{ref}$, the observed initial visual state (i.e., the reference
  image $\mI_\mathrm{ref}$), and the visual conditioning $\mathcal{C}({\tX})$ extracted from the predicted future
  video latents ${\tX} \in \mathbb{R}^{c \times (f/4) \times (h/8) \times (w/8)}$, formalizing an \emph{imagine-to-
  act} objective $\pi_{\gamma}=p(\mA \mid \mT_\mathrm{ref}, \mathcal{C}({\tX}), \mI_\mathrm{ref})$. Mirroring the
  video branch, the action chunk follows an independent flow-matching formulation: $\mA^{t'}=t'\mA^{1}+(1-t')\epsilon$, where the action timestep $t' \sim \mathcal{U}[0, 1]$ is sampled independently
  from the video timestep $t$. The $\phi$-parameterized action velocity expert $\mathcal{A}:\mathbb{R}
  ^{H \times d_a}\rightarrow\mathbb{R}^{H \times d_a}$ is trained by minimizing:
\begin{equation}\label{eq_act_loss}
\mathcal{L}_{\mathrm{act}} \;=\; \mathbb{E}_{t', \mA, \tX}\Big[\big\|\mathcal{A}(\mA^{t'}, t', \mathcal{C}({\tX}), \mI_\mathrm{ref}, \mT_\mathrm{ref};\phi) - (\mA^{1}-\epsilon)\big\|_2^2\Big].
\end{equation}
The overall training objective combines both experts via a weighted sum:
\begin{equation}\label{eq_video_action}
\mathcal{L} \;=\; \lambda_\mathrm{video}\mathcal{L}_{\mathrm{video}} \;+\; \lambda_\mathrm{act}\mathcal{L}_{\mathrm{act}}.
\end{equation}
where $\lambda_\mathrm{act}$  and $\lambda_\mathrm{video}$ are loss weights and the full policy is composed from $\gamma=\{\phi,\theta\}$. During the inference, the model performs chunk-level action generation through the same denoising process used for $\mathcal{V}$, thereby grounding control in the \emph{imagination} prior inherited from video generalists. 

Existing WAMs span a spectrum of visual contexts $\mathcal{C}(\mathbf{X})$ provided to the action policy: \emph{full-imagination} models~\citep{pai2025mimic,ye2026dreamzero} condition actions on complete predicted video clips ($\mathcal{C}(\mathbf{X})=\mathbf{X}_{1:f}$), whereas \emph{zero-imagination} models~\citep{yuan2026fastwam} omit generated future latents entirely ($\mathcal{C}(\mathbf{X})=\emptyset$). This presents a clear trade-off: \emph{full-imagination} offers dense spatial guidance but incurs substantial latency and video artifacts, whereas \emph{zero-imagination} enables real-time execution but lacks the visual foresight required for long-horizon tasks.



\section{ProWAM: Long-Horizon Control with Progressive Sub-goals}\label{sec_method}



\begin{figure*}[t]
\vspace{-4mm}
\centering
\includegraphics[width=\textwidth]{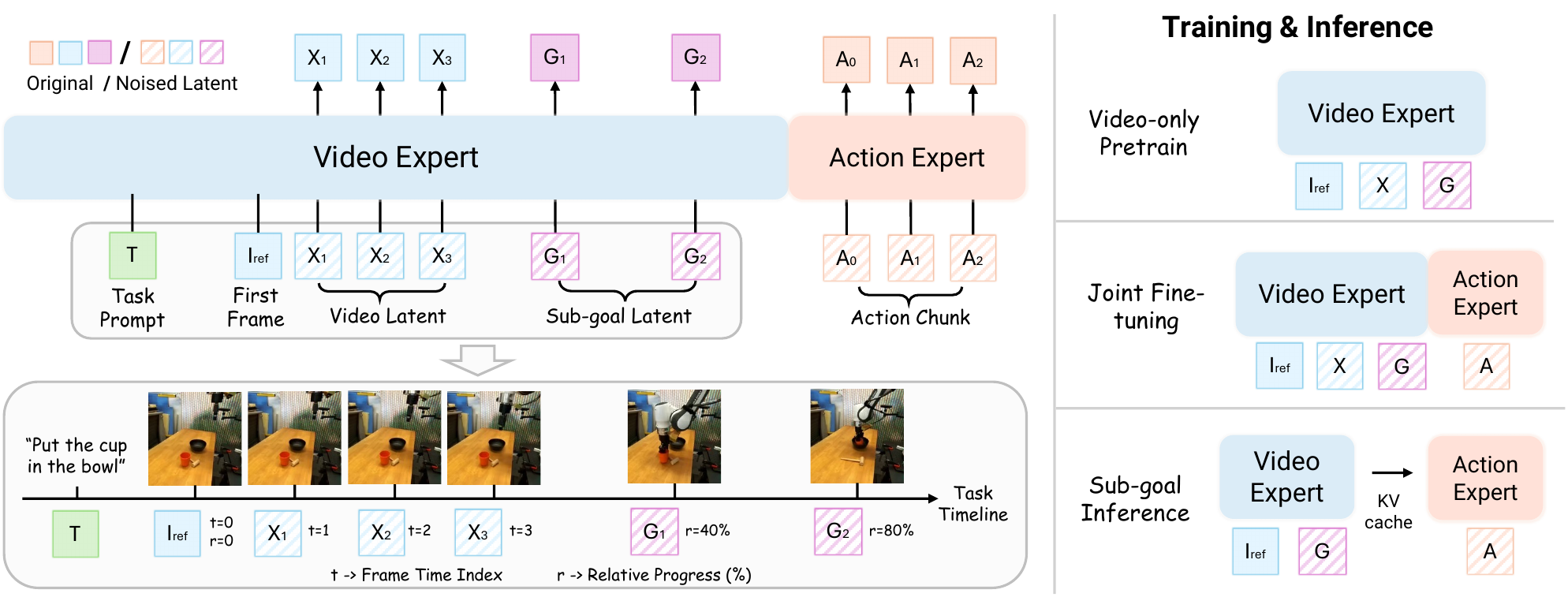}
\vspace{-5mm}
  \caption{\textbf{Overview of ProWAM.}
  \textbf{Left: Training architecture.}
  Given a task instruction $\mathbf{T}$ and reference observation
  $\mathbf{I}_{\mathrm{ref}}$, ProWAM jointly denoises future-video latents
  $\mathbf{X}$, progressive sub-goal latents $\mathbf{G}$, and action chunks
  $\mathbf{A}$. The bottom timeline aligns these latent variables with physical
  execution, with sub-goals providing visual targets at specified relative
  progress levels.
  \textbf{Right: Training and inference pipeline.}
  ProWAM employs two-stage training:
  (1) \emph{Video-only Pretraining}, where the video expert learns visual
  dynamics and progressive sub-goal prediction from action-free videos; and
  (2) \emph{Joint Fine-tuning}, where the video and action experts are jointly
  optimized on robot demonstrations, enabling the action expert to condition
  action generation on predicted sub-goals.
  At inference, the video expert predicts sparse sub-goal features instead of a
  dense video rollout. These features are cached and reused during action
  denoising to avoid repeated video-expert computation.}
\label{fig:overview}
\end{figure*}

\paragraph{\textbf{Beyond a Single Goal Guidance.}} To enable explicit visual planning in WAM, a natural baseline is \emph{final-goal prediction} (FGP)\footnote{Our pilot study finds that \emph{self-generated final goal} improves baselines in most settings (Appendix~\ref{sec_app_pilot}).}. However, compressing an entire episode into a solitary final frame specifies where the task terminates but provides no guidance on how to get there, leaving the intermediate execution path unspecified.
Inspired by hierarchical planning paradigms that decompose extended horizons into structured sub-states/goals~\citep{li2023hierarchical,luo2026generative}, we introduce a \emph{progressive sub-goal signal} directly into the WAM architecture. Specifically, we generalize FGP into a \emph{progressive sequence of sparse $k$ sub-goals} that   predicts an ordered sequence of sparse sub-goals at selected relative future positions, yielding our \emph{progressive world action model} (\textbf{ProWAM}). Crucially, while classical hierarchical planners~\citep{intelligence2026pi07steerablegeneralistrobotic,shou2026halo} depend on dedicated, multi-stage goal-generation modules, ProWAM generates the entire sub-goal trajectory natively and injects their features directly to the action expert (as shown in Figure~\ref{fig:overview}).

  \paragraph{\textbf{Progressive Sub-goal Parameterization.}}
  We parameterize intermediate sub-goals using a normalized \emph{progress
  coordinate} $r\in[0,1]$, defined as the relative temporal position along the
  remaining demonstration trajectory from the current observation ($r=0$) to
  its endpoint ($r=1$).  This normalization provides a consistent temporal progress scale across
  trajectories of different lengths. Given $k$
  monotonically increasing progress levels
  $0<r_1<\dots<r_k\leq1$, the $i$-th sub-goal corresponds to the frame at
  \begin{equation}\label{eq:progress_index}
  \tau(r_i)=\tau_0+
  \left\lfloor r_i\big(\tau_{\mathrm{end}}-\tau_0\big)\right\rceil,
  \qquad i=1,\ldots,k,
  \end{equation}
  where $\tau_0$ and $\tau_{\mathrm{end}}$ denote the current-frame and
  demonstration-endpoint indices, respectively, and $\lfloor\cdot\rceil$ denotes
  nearest-integer rounding. Each selected frame is independently encoded as
  $\tilde{G}_i=\mathcal{E}(V_{\tau(r_i)})
  \in\mathbb{R}^{c\times1\times(h/8)\times(w/8)}$. The resulting sequence spans selected future states, from near-term goal at
  smaller $r_i$ to longer goal as $r_i \rightarrow 1$ ($r_1=1, k=1$ in FGP). During training, we sample
  $k\sim\mathcal{U}\{1,\dots,K_{\max}\}$, enabling flexible sub-goal learning strategy.   At inference, $\tG$ is predicted relative to the latest observation,
  which serves as $r=0$ at each replanning step.

To incorporate progressive sub-goal guidance directly into the WAM architecture, we append the progressive sub-goal sequence to the video latent sequence, yielding:
\begin{equation}\label{eq:layout}
\tZ \;=\; \big[\;\underbrace{\tX_{0}}_{\text{current observation}}
\;\big\|\; 
\underbrace{\tX_{1},\dots,\tX_{f/4-1}}_{\text{clip transitions}}
\;\big\|\; \underbrace{\tG_1,\dots,\tG_k}_{\text{progressive sub-goals}}\;\big],
\end{equation}
where $\tX$ denotes the $f/4$ video latent frames, with $\tX_0$ representing the initial observation. During training, $\tX_0$ serves as a clean observation latent at every denoising timestep, whereas the transition clip and sub-goal latents follow the same flow-matching process. This formulation preserves full-sequence video prediction while adding the guidance of explicit progress-conditioned sub-goals.

\paragraph{\textbf{Progress-conditioned Modulation.}}
Sub-goal slots share the same network weights and temporal position encodings as the video latents. To specify the target temporal stage of each sub-goal, we additionally condition every slot on its progress coordinate $r_i \in [0, 1]$. Specifically, $r_i$ is mapped through a sinusoidal embedding followed by a \emph{multi-layer perceptron} (MLP) to yield an \emph{adaptive LayerNorm} (adaLN)~\citep{peebles2023scalable} modulation offset $\Delta_{\mathrm{mod}}(r_i)$. This offset is added directly to the standard flow-matching timestep modulation:
\begin{equation}\label{eq:progress_mod}
\text{mod}(\tG_i) \;=\; \text{mod}(t) \;+\; \Delta_{\mathrm{mod}}(r_i).
\end{equation}
This modulation anchors each sub-goal to its corresponding task stage relative to the reference observation. Trained under progress conditioning and varying sub-goal counts $k$, ProWAM natively generalizes to flexible sub-goal schedules at inference without architectural modifications.

{\textbf{Asymmetric Vision-Action Coupling.}} 
Following~\citet{bi2026motus,yuan2026fastwam}, ProWAM implements the video expert $\mathcal{V}$ and action expert $\mathcal{A}$ via a \emph{mixture-of-transformers} (MoT)~\citep{liang2024mixture} architecture with shared attention heads. This design enables $\mathcal{V}$ to retain its pretrained generative representations while $\mathcal{A}$ specializes in action generation by directly conditioning on predicted visual features. Information flow across the experts is governed by a structured visibility mask:

\begin{itemize}[label=$\bullet$, leftmargin=1.5em, itemsep=2pt, topsep=2pt]
\setlength{\parskip}{0pt}
\item \textbf{Sub-goal planning} {[$\tG \leftrightarrow \tG, \tG \to \tX_0$]}\footnote{The notation $A \to B$ ($A \not\to B$) indicates that queries in $A$ attend (do not attend) to keys/values in $B$, whereas $A \leftrightarrow A$ denotes bidirectional self-attention within $A$. Note that $\tX_0$ only attends to itself.}: $\tG$ attend bidirectionally among themselves and to the reference observation $\tX_0$. This joint attention enables intermediate milestones to share visual features, allowing information exchange across sub-goal slots  rather than $k$ isolated predictions.

\item \textbf{Transition conditioning} {[$\tX \leftrightarrow \tX, \tX \to \{\tX_0, \tG\}$]}: $\tX$ attend bidirectionally among themselves to maintain temporal coherence, while conditioning on $\tX_0$ and $\tG$ as guidance anchors. This encourages visual consistency between the generated video and the planned sub-goals.

\item \textbf{Asymmetric vision-action coupling} {[$\mA \leftrightarrow \mA, \mA \to \{\tX_0, \tG\}, \tX \not\to \mA$]}: $\mathcal{V}$ does not attend to action tokens ($\tX \not\to \mA$),  maintaining an action-independent visual forward pathway of $\mathcal{V}$. Conversely, action tokens attend bidirectionally among themselves and cross-attend to visual anchors $\{\tX_0, \tG\}$, grounding low-level action execution directly in the predicted sub-goals.
\end{itemize}

Thus, the joint policy $\pi_\gamma(\mA \mid \mT_\mathrm{ref}, \mathcal{C}({\tZ}), \mI_\mathrm{ref})$ (with parameters $\gamma = \{\phi, \theta\}$) is explicitly conditioned on the visual context chain anchored by the current observation:
\begin{equation}\label{eq:context}
\mathcal{C}(\tZ) \;=\; \{\tX_0\} \cup \{\tG_1,\dots,\tG_k\}.
\end{equation}

\paragraph{\textbf{Training Objective.}}
To jointly enable progressive visual imagination and downstream action control, ProWAM is trained via flow matching with independent timesteps for the video expert ($t \sim \mathcal{U}[0, 1]$) and action expert ($t' \sim \mathcal{U}[0, 1]$), where sub-goals $\tG$ share the video timestep $t$. The full objective combines three distinct velocity-regression terms:
\begin{equation}\label{eq:full_loss}
\mathcal{L}_\mathrm{final} \;=\; \lambda_{\mathrm{video}}\mathcal{L}_{\mathrm{video}} \;+\; \lambda_{\mathrm{sub}}\,\mathcal{L}_{\mathrm{sub}} \;+\; \lambda_{\mathrm{act}}\,\mathcal{L}_{\mathrm{act}},
\end{equation}
where $\mathcal{L}_{\mathrm{video}}$ and $\mathcal{L}_{\mathrm{sub}}$ supervise flow-matching velocity predictions for transition latents $\tX$ and sub-goal latents $\tG$, respectively. Similarly, $\mathcal{L}_{\mathrm{act}}$ trains the action expert to predict velocities for action chunks $\mA$, conditioned on the visual context $\mathcal{C}(\tZ)$, reference observation $\mI_{\mathrm{ref}}$, and task instruction $\mT_{\mathrm{ref}}$. We optimize this objective across 2 distinct phases to preserve generative priors while learning control:
In \textbf{Stage 1} (Video-only Training), we pretrain the video expert on action-free video corpora using $\lambda_{\mathrm{video}}\mathcal{L}_{\mathrm{video}} + \lambda_{\mathrm{sub}}\mathcal{L}_{\mathrm{sub}}$ to learn visual dynamics and progress-conditioned sub-goal prediction. In \textbf{Stage 2} (Joint Video-Action Fine-tuning), we jointly fine-tune both experts on robot demonstrations using $\mathcal{L}_{\mathrm{final}}$, grounding action execution directly in the planned visual sub-goals.
Consequently, ProWAM can leverage vast, action-free video corpora for scalable pre-training, drastically improving sample efficiency when fine-tuning on limited, expensive robot demonstrations.

\paragraph{\textbf{Efficient Inference with Sub-goal Caching.}}

Rather than performing burdensome joint video-action denoising that advances the video and action branches synchronously across $T$ steps~\citep{ye2026dreamzero},
we empirically find that sub-goal features obtained from a single video-expert evaluation support comparable policy performance to full joint denoising (Appendix~\ref{sec_app_downstream}). At each replanning step, we evaluate $\mathcal{V}$ once to predict the sub-goals and cache their key--value features. The action head then reuses this fixed cache throughout $T$ action-denoising steps. Upon receiving a new observation, ProWAM recomputes the sub-goals and refreshes the cache. This reduces video-expert forward passes from $T$ to $1$ ($\approx10\times$ fewer video-expert evaluations), achieving inference cost comparable to prior cached WAMs~\citep{pai2025mimic,yuan2026fastwam,zhang2026imagewam}.



\section{Experiments}
\label{sec_exo}
\subsection{Experimental Settings}
Following the evaluation protocols in concurrent works~\citep{25pi0.5,bu2025univla,yuan2026fastwam,beingbeyond2025beingh0}, we adopt several prevailing benchmarks for \emph{in-domain} (ID) and \emph{out-of-domain} (OOD) simulation evaluation, including  LIBERO~\citep{LIBERO23} \& LIBERO-Plus~\citep{fei25libero-plus}, RoboTwin~\citep{chen2025robotwin}, and RoboCasa365~\citep{nasiriany2024robocasa}, alongside zero-shot real-world robot deployments on Franka. We strictly follow the standard evaluation metrics tailored for each benchmark and report the average success rate for each task. Our ProWAM adopts the Wan2.2-TI2V-5B~\citep{wan2025wan} as the video backbone. Since our sub-goal-conditioned prediction relies on a strong visual world prior, we study two paradigms that differ in the strength of that prior: (i) \textbf{ProWAM-Base}, initialized directly from the Wan base model (w/o \textbf{Stage-1}). (ii) \textbf{ProWAM}, initialized from a video-only paradigm pretrained on large-scale robot and egocentric data (see Appendix~\ref{sec_appendix_pre} for details). The video branch still generates
sub-goals as an auxiliary co-training objective, so the shared representation
remains sub-goal-aware. At inference, the action expert does not require fully denoised sub-goal latents but conditions on sub-goal key-value features cached from a single video-expert evaluation, yielding a simpler
and faster policy. Additional experimental details are deferred to Appendix~\ref{sec_app_downstream} \& \ref{sec_app_Real Robotic Experiment}.



\subsection{Simulated Policy Evaluation}\label{sec_app_Simulated Policy Evaluation}

\begin{table*}[ht]
\centering
\setlength{\tabcolsep}{2pt}
\renewcommand{\arraystretch}{1.08}

\begin{minipage}[t]{0.49\textwidth}
\centering
\caption{\textbf{Success rate (\%) on LIBERO-series.} 
Plus* is strictly used for zero-shot evaluation.}
\vspace{-2mm}
\label{tab:libero}

\small
\begin{tabular*}{\linewidth}{@{\extracolsep{\fill}}lcc@{}}
  \toprule
  Method & LIBERO & Plus* \\
  \midrule
  \rowcolor{gray!12}
  \multicolumn{3}{l}{\emph{Vision-Language-Action Models (VLA)}} \\
  $\pi_0$~\citep{24pi0}                    & 94.1 & 53.6 \\
  $\pi_0$-Fast~\citep{pertsch2025fast}     & 85.5 & 61.6 \\
  StarVLA~\citep{ye2026starvla}            & 96.5 & 77.0 \\
  $\pi_{0.5}$~\citep{25pi0.5}              & 96.9 & 77.4 \\
  ABot-M0~\citep{yang2026abot}             & 98.6 & 80.5 \\
  \rowcolor{gray!12}
  \multicolumn{3}{l}{\emph{World Action Models (WAM)}} \\
  VLA-JEPA~\citep{sun2026vla}              & 96.1 & 79.5 \\
  Fast-WAM~\citep{yuan2026fastwam}         & 97.6 & 51.5 \\
  Image-WAM~\citep{zhang2026imagewam}      & 98.4 & 83.1 \\
  Being-H0.7~\citep{luo2026being}          & 99.2 & 82.1 \\
  \midrule
  \textbf{ProWAM-Base}                     & \textbf{99.2} & 80.3 \\
  \textbf{ProWAM}                          & 98.8 & \textbf{85.8} \\
  \bottomrule
\end{tabular*}
\end{minipage}%
\hfill 
\begin{minipage}[t]{0.49\textwidth}
\centering
\caption{\textbf{Success rate (\%) on RoboTwin.} 
Random* is used for OOD evaluation.}
\vspace{-2mm}
\label{tab:robotwin}

\small
\begin{tabular*}{\linewidth}{@{\extracolsep{\fill}}lcc@{}}
  \toprule
  Method & Clean & Random* \\
  \midrule
  \rowcolor{gray!12}
  \multicolumn{3}{l}{\emph{Vision-Language-Action Models (VLA)}} \\
  DP~\citep{chi2025diffusion}              & 28.0 & 0.6 \\
  RDT~\citep{liu2024rdt}                   & 34.5 & 13.7 \\
  $\pi_0$~\citep{24pi0}                    & 46.4 & 16.3 \\
  UP-VLA~\citep{upvla25}                   & 52.9 & 15.2 \\
  \rowcolor{gray!12}
  \multicolumn{3}{l}{\emph{World Action Models (WAM)}} \\
  Fast-WAM~\citep{yuan2026fastwam}         & 71.9 & 6.3 \\
  BagelVLA~\citep{hu2026bagelvla}          & 75.3 & 20.5 \\
  HALO~\citep{shou2026halo}                & 80.5 & 26.4 \\
  Image-WAM~\citep{zhang2026imagewam}      & 85.0 & 37.6 \\
  MV-WAM~\citep{chen2026mv}                & 84.0 & 55.7 \\
  \midrule
  \textbf{ProWAM-Base}                     & \textbf{87.6} & 67.4 \\
  \textbf{ProWAM}                          & 85.8 & \textbf{75.7} \\
  \bottomrule
\end{tabular*}
\end{minipage}
\vspace{-3mm}
\end{table*}

\paragraph{\textbf{LIBERO \& LIBERO-Plus}.} Table~\ref{tab:libero} provides a comprehensive quantitative comparison between two variants of our model and prior baselines across the LIBERO benchmark suite. Notably, both of our models achieve near-saturation performance on the in-domain LIBERO tasks ($\geq 98.6\%$). As a result, the primary discriminative signal for evaluating generalizability shifts to the challenging zero-shot LIBERO-Plus evaluation setting. Under this regime, our ProWAM achieves a new \emph{state-of-the-art} (SOTA) performance of $85.8\%$, markedly outperforming previous strong WAM and VLA baselines. Furthermore, ProWAM-Base reveals a sharp performance drop ($5.5$ lower) on zero-shot tasks, highlighting the importance of sub-goal pretraining for OOD robustness.


  \paragraph{\textbf{RoboTwin}.}Table~\ref{tab:robotwin} reports results on RoboTwin, where vision-action
  fine-tuning uses \emph{only clean demonstrations} and evaluated under both in-distribution and randomized 
  conditions. ProWAM achieves $75.7\%$ success under scene randomization,
  outperforming the strongest prior WAM by \textbf{+20.0}. Removing
  video pretraining reduces randomized OOD success from $75.7\%$ to $67.4\%$,
  while clean-scene performance remains comparable ($85.8\%$ versus $87.6\%$).
  These results indicate that video pretraining primarily improves robustness to
  visual domain shifts.


\begin{table}[h]
  \centering
\caption{\textbf{Closed-loop performance on Public Methods in RoboCasa365}. Full results are reported in Appendix~\ref{sec_app_Simulated Policy Evaluation}. 
}
  \label{tab:robocasa_part}
  \resizebox{\linewidth}{!}{
  \begin{tabular}{clcccc}
    \toprule
    Ranking & Method & Overall & Atomic-Seen & Composite-Seen & Composite-Unseen \\
    \midrule
    1 & Paimon-0           & 58.1 & 79.3 & 55.8 & 36.7 \\
    2 & Xiaomi-Robotics-1~\citep{team2026xiaomi}    & 57.4 & 80.2 & 57.1 & 32.1 \\
    3 & Phasor-m7    & 56.8 & 78.7 & 57.4 & 31.5 \\
    \rowcolor{gray!12} 
    4 & \textbf{ProWAM (Ours)} & \textbf{48.1} & \textbf{76.0} & \textbf{46.5} & \textbf{18.2}
    \\
    5 & ABot-M0.6~\citep{chen2026abot}            & 46.6 & 79.4 & 48.3 & 7.9
    \end{tabular}}
    \end{table}

{\textbf{RoboCasa365.}} Table~\ref{tab:robocasa_part} reports results on RoboCasa365. 
ProWAM achieves a \textbf{48.1\%} overall success rate and ranks \textbf{second overall} among the evaluated methods. It also delivers competitive performance on long-horizon composite tasks. Notably, ProWAM achieves the \textbf{second-highest success rate} on the challenging long horizon \emph{Composite-Unseen} split, substantially outperforming prior world-model baselines. These results support the effectiveness of progressive visual sub-goals for long-horizon execution and generalization to held-out task compositions.

\subsection{Zero-shot Real-World Evaluation} 
\label{sec_app:exp_setup_real}

\paragraph{\textbf{Evaluation protocol \& Results}}
\begin{wraptable}{r}{0.38\linewidth}
    \caption{\textbf{Real-world performance.}}
  \centering
  \small
  \begin{tabular}{lcc}
    \toprule
    \textbf{Method} & \textbf{Size} &\textbf{SR (\%)} \\
    \midrule
    $\pi_{0.5}$   & 3B & 50.0 \\
    DreamZero & 14B & 55.0 \\
    \midrule
    \textbf{ProWAM (Ours)} & 5B & \textbf{70.0} \\
    \bottomrule
  \end{tabular}
  \label{tab:sim_results}
\end{wraptable}
To evaluate true out-of-the-box generalization without collecting
environment-specific demonstrations or retraining baselines, we adopt
DROID~\citep{khazatsky2024droid} as the unified real-robot training dataset. Table~\ref{tab:sim_results} summarizes the real-world evaluation across the $4$ physical tasks, which span basic \emph{pick-and-place}, \emph{long-horizon multi-object assembly}, \emph{contact-rich placement}, and \emph{spatially grounded placement} (full task details are given in Appendix~\ref{sec_app_Real Robotic Experiment}). ProWAM achieves an average success rate of \textbf{70.0\%}. Notably, ProWAM outperforms DreamZero by a \textbf{+15.0} absolute margin, exhibiting particular advantages in spatial placement, where it reaches \textbf{100.0\%} success against the baselines. These results demonstrate that ProWAM enables direct zero-shot generalization without requiring domain-specific fine-tuning.

\begin{figure*}[ht]
    \centering
    \includegraphics[width=\textwidth]{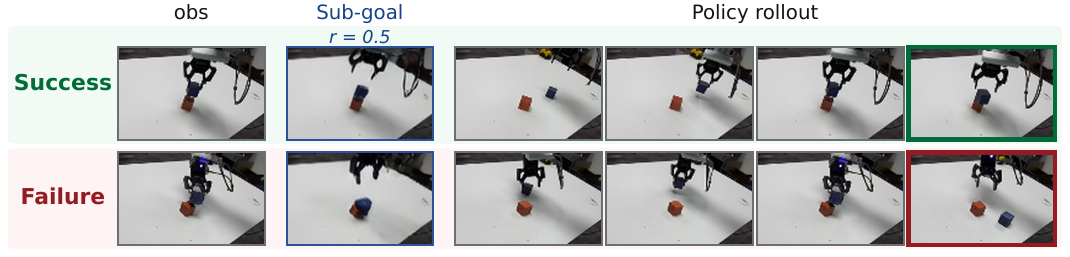}
    \vspace{-7mm}
  \caption{\textbf{Failure case on ``Cube Stacking.''}
  The failed sub-goal (\emph{bottom}) exhibits more ambiguous contact geometry
  than the successful one (\emph{top}), and the
  gripper subsequently opens prematurely.}
    \label{fig:real_failure_case}
    \vspace{-2mm}
\end{figure*}

\paragraph{\textbf{Failure Case Analysis.}}
Figure~\ref{fig:real_failure_case} compares sub-goals generated immediately before release in successful and failed ``Cube Stacking'' rollouts. In the successful rollout, the sub-goal depicts a centered placement with a clear contact boundary. In the failure case, the two cubes appear blurred together, obscuring the intended contact geometry, then the subsequent execution releases the cube off-center and drops it. This qualitative correspondence suggests that policy execution can tolerate texture imperfections but may remain sensitive to errors in task-relevant geometry.

\subsection{Ablation Studies}

We conduct ablation studies on RoboTwin, whose moderate success rates offer a sufficient dynamic range to differentiate design choices. In contrast, LIBERO's near-saturated in-distribution SR ($\sim\!99\%$) obscures performance differences within run-to-run noise, making it far less discriminative.

\begin{figure}[ht] 
  \centering
  \begin{minipage}[c]{0.32\linewidth}
    \centering
    \includegraphics[width=\linewidth]{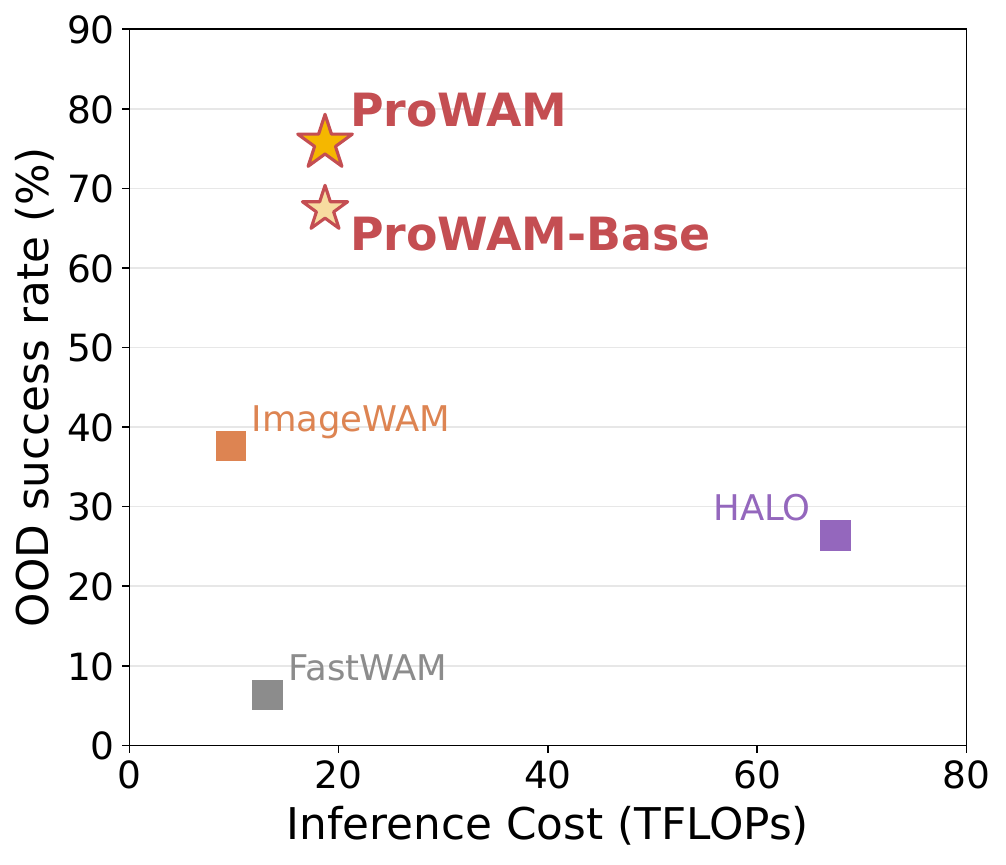} 
    \par\vspace{-2pt}
    {\footnotesize (a) \textbf{Efficiency vs.\ Performance.}}
  \end{minipage}%
  \hfill
  \begin{minipage}[c]{0.32\linewidth}
    \centering
    \includegraphics[width=0.97\linewidth]{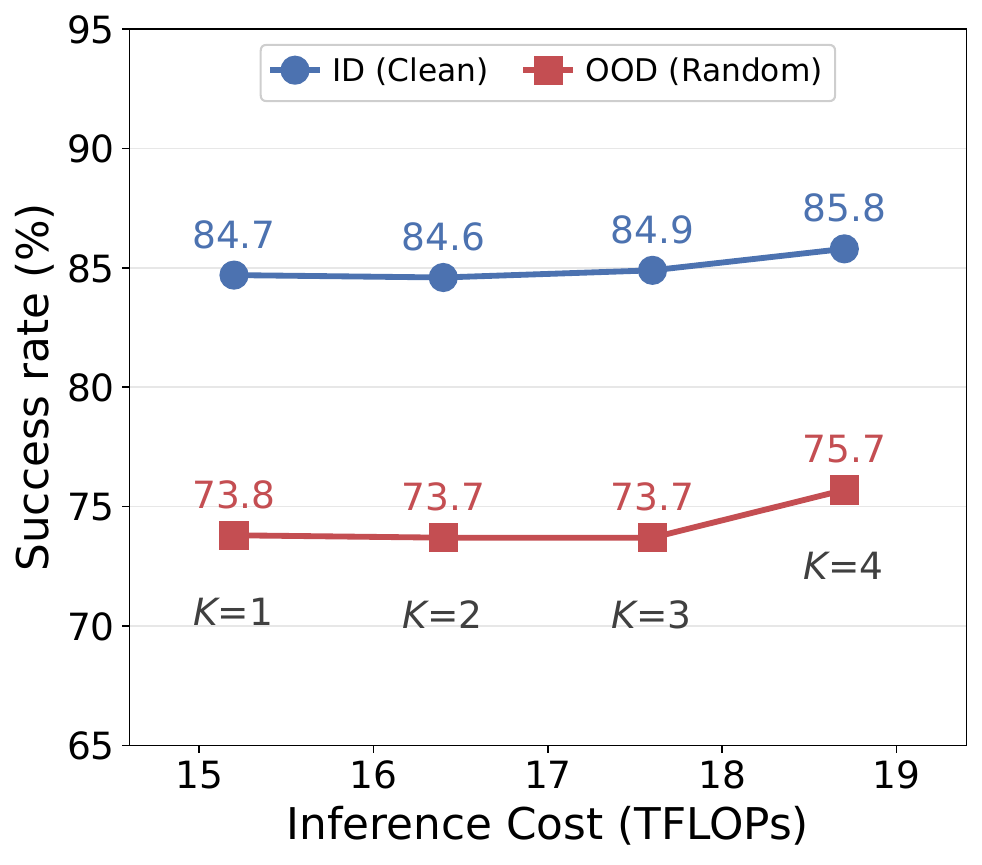} 
    \par\vspace{-2pt}
    {\footnotesize (b) \textbf{Effect of Sub-goal Number.}}
  \end{minipage}%
  \hfill
  \begin{minipage}[c]{0.32\linewidth}
    \centering
    \includegraphics[width=\linewidth]{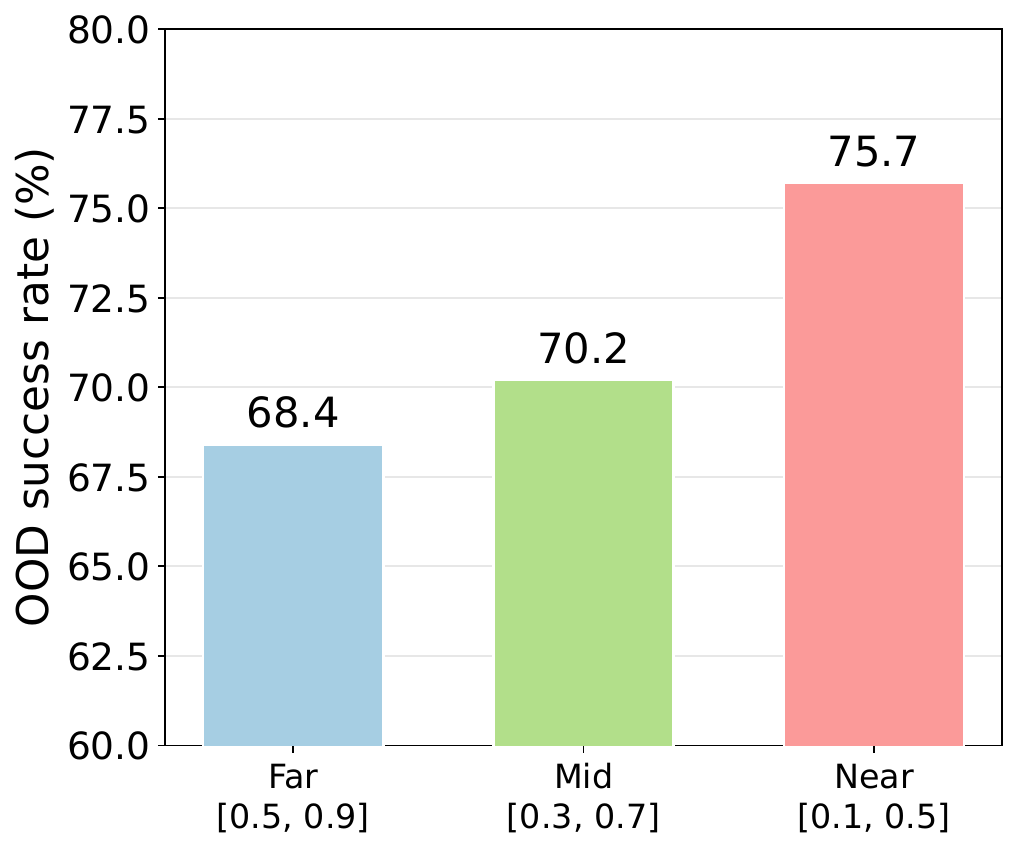} 
    \par\vspace{-2pt}
    {\footnotesize (c) \textbf{Effect of Sub-goal Horizon.}}
  \end{minipage}
  \vspace{-1mm}
  \caption{\textbf{Efficiency and sub-goal ablations.}
  \textbf{(a)} OOD success versus inference computation across WAMs.
  \textbf{(b)} Performance and computation across different sub-goal counts, where
$k=4$ yields peak OOD performance with efficient inference.
  \textbf{(c)} Performance across \texttt{Near}, \texttt{Mid}, and \texttt{Far}
  progress schedules. \texttt{Near} achieves the best OOD result while ID
  performance remains stable.}
  \label{fig:ablation_effiency_subgoal}
\end{figure}

\begin{figure*}[ht]
\vspace{-3mm}
    \centering
    \includegraphics[width=\linewidth]{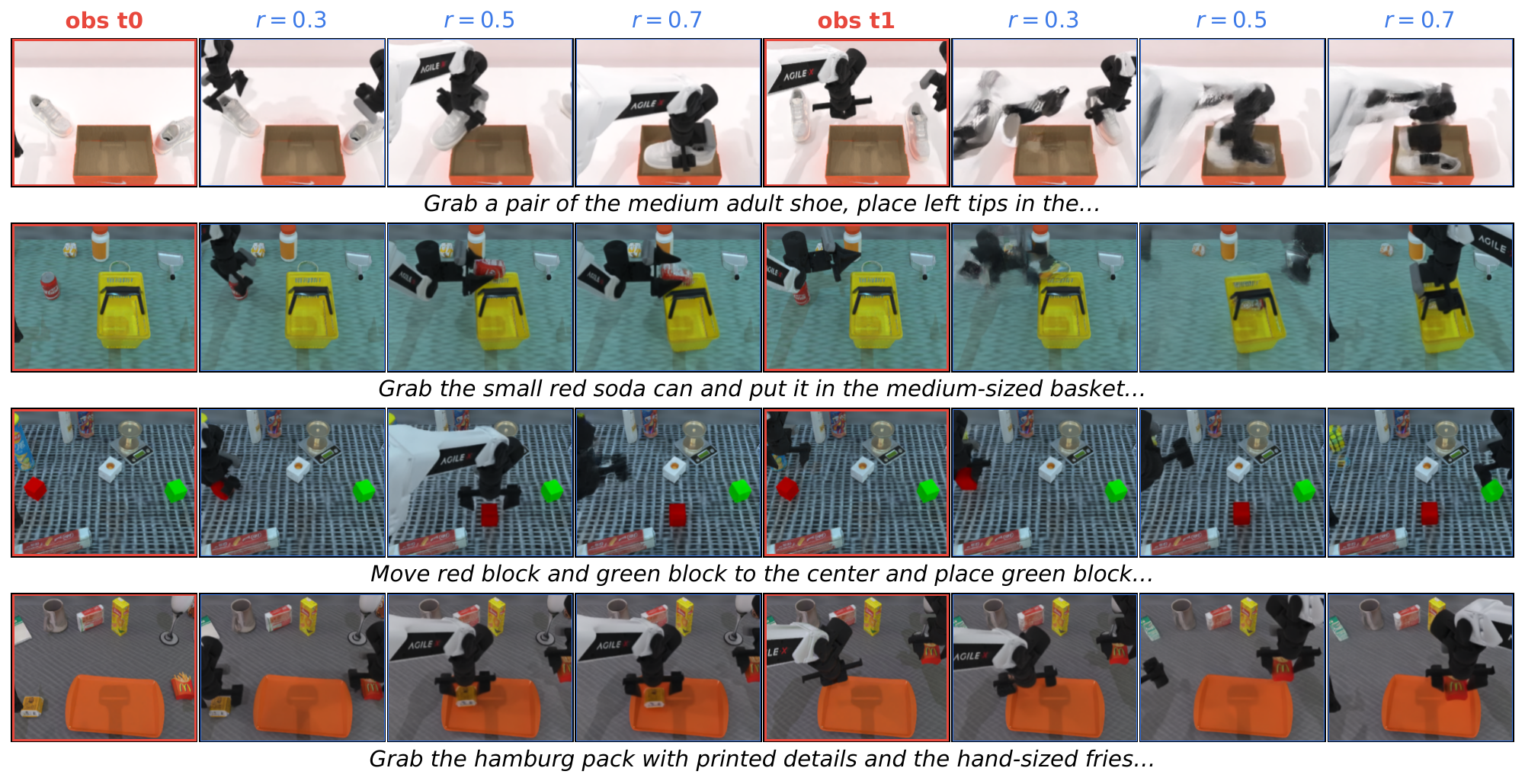}
    \vspace{-5mm}
  \caption{\textbf{Closed-loop sub-goal visualization.}
  For each observation at \textcolor{red}{$t_0$ and $t_1$}, the model generates sub-goals in a single pass at progress levels
  \textcolor{blue}{$r\!\in\!\{0.3,0.5,0.7\}$} to guide action generation.}
    \label{fig:subgoal_viz}
\end{figure*}

\paragraph{\textbf{Efficiency \& Sub-goals Number.}}
  To evaluate computational efficiency, we benchmark inference latency and FLOPs
  for a $16$-action chunk on a single NVIDIA H100 GPU. Compared with naive joint
  denoising ($115.5$~TFLOPs, $581$~ms), our KV-cache strategy reduces computation
  to $18.7$~TFLOPs ($84\%$ reduction) and latency to $313$~ms ($46\%$ reduction),
  while achieving 3.3 percentage points higher OOD success in this evaluation.
  As shown in Figure~\ref{fig:ablation_effiency_subgoal}(a), ProWAM achieves a
  favorable trade-off between computational efficiency and robustness: while
  in-distribution performance remains similar across methods, ProWAM reaches
  $75.7\%$ OOD success under domain randomization, outperforming other methods at
  comparable or lower computational cost.

  Furthermore, $k$ serves as an inference-time hyperparameter: a single model
  trained with $k\in\{1,\ldots,4\}$ supports all four settings
  (Figure~\ref{fig:ablation_effiency_subgoal}(b)). In-domain and OOD performance remains
  stable across $k \in [1,3]$, and reaches its highest value of $75.7\%$ at $k=4$. Because all
  sub-goals are predicted in one cached forward pass, increasing $k$ from $1$ to
  $4$ adds only $3.5$~TFLOPs. Even at $k=1$, ProWAM achieves $73.8\%$ OOD success
  at $15.2$~TFLOPs, substantially outperforming FastWAM's $6.3\%$ at a comparable
  computational cost of $13.2$~TFLOPs.

\paragraph{\textbf{Sub-goal Planning Horizon.}}
We evaluate three progress schedules under identical training settings: \texttt{Near} ($r\in[0.1,0.5]$), \texttt{Mid} ($r\in[0.3,0.7]$), and \texttt{Far} ($r\in[0.5,0.9]$) (Figure~\ref{fig:ablation_effiency_subgoal}(c)). Shorter horizons yield monotonic OOD gains, rising from $68.4\%$ (\texttt{Far}) to $70.2\%$ (\texttt{Mid}) and $75.7\%$ (\texttt{Near}), while ID performance remains stable around $86\%$. This $+7.3$ improvement suggests that nearer targets offer tighter visual grounding for action generation, whereas distant goals leave larger unguided execution gaps. We thus adopt the \texttt{Near} schedule by default.


\paragraph{\textbf{Sub-goal Generation.}} 
Figure~\ref{fig:subgoal_viz} visualizes sub-goals generated in a single cached forward pass during closed-loop execution. First, the predicted states advance with increasing $r$ and adapt to the latest observation at each replanning step. Second, while single-step generation introduces minor visual artifacts (e.g., blurred textures), sub-goals preserve salient scene structure and task progression, providing geometric guidance for action generation without requiring pixel-perfect reconstruction.


\section{Related Work}\label{sec_related}

\paragraph{\textbf{World Action Models: Conditioning Control on Future Imagination.}}
Generative video priors~\citep{blattmann2023svd,wan2025wan} have spurred the development of \emph{world action models} (WAMs), which leverage visual foresight to condition robotic policies~\citep{wang2026world}. Early frameworks adopted cascaded pipelines, first predicting future visual states and subsequently recovering actions via inverse dynamics or learned policies~\citep{du2023video,bharadhwaj2024gen2act}. Modern WAMs tightly unify predictive modeling and action generation within joint diffusion objectives and shared architectures~\citep{kim2026cosmos,bi2026motus}. These approaches form an \emph{imagination spectrum} according to the future visual context used for action generation. \emph{Full-imagination} WAMs~\citep{pai2025mimic,ye2026dreamzero,li2026lingbotva} condition actions on dense generated video rollouts, providing detailed short-horizon guidance at high computational cost. In contrast, \emph{zero-imagination} WAMs~\citep{yuan2026fastwam,ye2026gigaworldpolicy} use no predicted future visual states during inference. Between them, \citet{zhang2026imagewam} conditions action generation on a single future target. ProWAM instead predicts a variable-length sequence of future visual
  states indexed by relative progress, providing guidance at multiple selected
  stages of the remaining trajectory.

\paragraph{\textbf{Vision-Language-Action Models: Leveraging Foundation Knowledge.}}
\emph{Vision-language-action} (VLA) models transfer semantic priors from multimodal pretraining~\citep{bai2025qwen3} to robotic action generation by mapping task instructions and visual observations to actions~\citep{ma2026vlasurvey}. Building on generalist models~\citep{kim2024openvla,team2024octo,liu2024rdt}, existing VLAs represent actions using either discrete tokens~\citep{RT223,pertsch2025fast} or continuous diffusion and flow-matching objectives~\citep{24pi0,liu2024rdt,chi2025diffusion}. For long-horizon tasks, recent hierarchical VLAs additionally introduce textual sub-instructions, multimodal reasoning traces, or visual sub-goals~\citep{RTH24,intelligence2026pi07steerablegeneralistrobotic,shou2026halo,chen2026vistaenhancingvisualconditioning}. While effective, many of these approaches rely on \emph{decoupled} high-level modules or multi-stage inference pipelines, increasing computational complexity and inducing cascading error propagation across stages. ProWAM instead \emph{natively} predicts progressive visual sub-goals within its video-generation pathway and directly couples them to action generation inside a \emph{unified} WAM.

\paragraph{\textbf{Integrating Intermediate Foresight into End-to-End Control.}}
Directly executing long-horizon manipulation through unconditioned action policies often suffers from compounding trajectory drift. Decomposing complex tasks into intermediate visual sub-goals provides structured foresight, simplifying long-horizon control into manageable visual milestones~\citep{li2023hierarchical,xie2026sva}. Existing approaches incorporate such foresight either through standalone goal-generation modules~\citep{shou2026halo} or via online model-based trajectory search over candidate rollouts~\citep{neary2025vlaps,guo2025vlareasoner}. Nevertheless, modular goal generators remain vulnerable to error propagation across decoupled stages, while online trajectory search introduces prohibitive computational latency for real-time control. ProWAM instead integrates progress-conditioned sub-goal prediction and action generation through shared attention, reusing cached sub-goal features during iterative action denoising without requiring a separate planner or online trajectory search.

\section{Conclusion}

  In this paper, we introduced ProWAM, a world action model that integrates visual
  planning and physical control within a unified generative framework. Motivated
  by hierarchical policies that decompose long-horizon tasks into intermediate
  sub-goal states, we incorporate progressive visual sub-goal prediction directly
  into the world-modeling process without requiring a separate high-level planner.
  Specifically, ProWAM uses a generative video expert to predict visual sub-goals
  indexed by relative temporal progress and couples their features to low-level
  action generation, without auxiliary modules or test-time search. Moreover, the unified
  architecture supports pretraining on large-scale action-free videos followed by
  joint vision-action fine-tuning, allowing visual dynamics learned
  without action labels to inform downstream action generation. Evaluations across simulation benchmarks and
  four physical manipulation tasks show strong performance under the evaluated
  in-domain, randomized, and held-out long horizon settings. Overall, our results support
  progressive visual prediction as an effective interface between visual
  imagination and robotic control.


\clearpage
\newpage
\bibliographystyle{assets/plainnat}
\bibliography{paper}

\clearpage
\newpage

\clearpage
\appendix

\section*{Appendix Overview} 

\startcontents[appendix]
\printcontents[appendix]{l}{1}{\setcounter{tocdepth}{2}}

\vspace{8mm}
\hrule
\vspace{8mm}

\noindent In this supplementary document, we provide pilot study analyses, additional experimental details and analysis, failure cases, and extended discussions referenced in the main text.

\clearpage 

\section{Pilot Experiment}\label{sec_app_pilot}

\begin{figure*}[ht]
  \centering
  \begin{subfigure}[b]{0.68\textwidth}
    \centering
    \includegraphics[width=\textwidth]{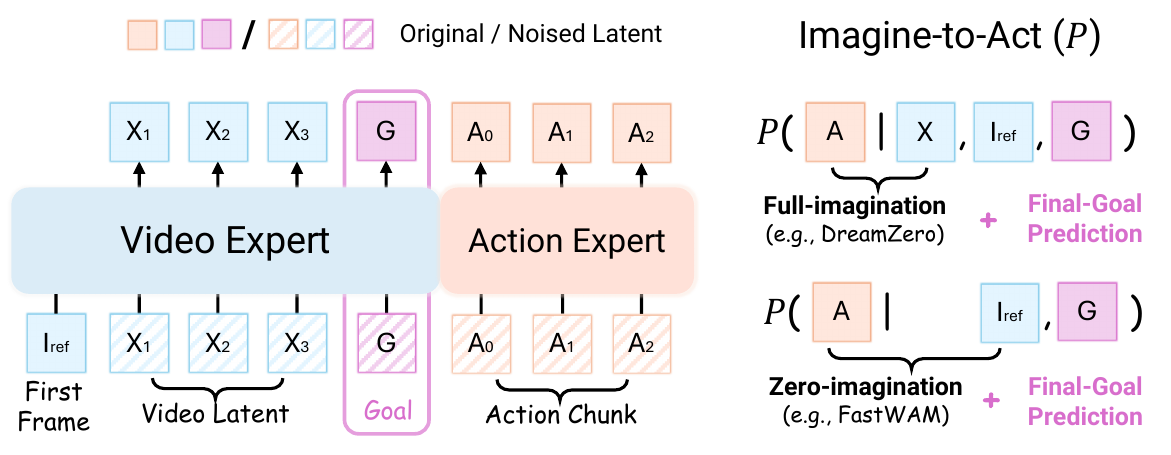}
  \caption{\textbf{The joint WAM architecture augmented with FGP.}}
    \label{fig:pilot}
  \end{subfigure}
  \hspace{2mm}
  \begin{subfigure}[b]{0.29\textwidth}
    \centering
    \includegraphics[width=\textwidth]{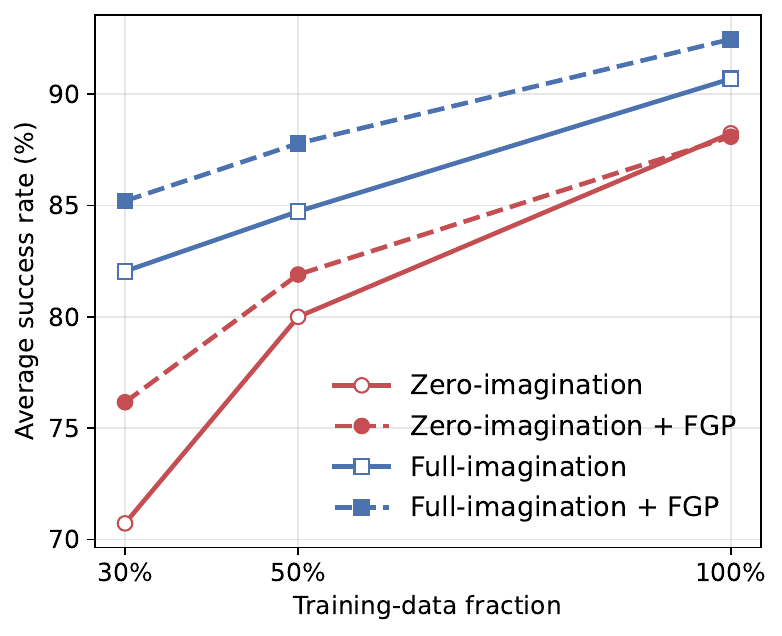}
\caption{\textbf{Pilot Evaluation.} FGP performance across three datasets.}
    \label{fig:pilot_plot}
  \end{subfigure}
  

\caption{\textbf{(a)} \textbf{Standard WAMs instantiate the \emph{imagine-then-act} formulation via two effective regimes}: \emph{full-imagination} (e.g., DreamZero~\citep{ye2026dreamzero}), which conditions on the full video trajectory $X$, and \emph{zero-imagination} (e.g., FastWAM~\citep{yuan2026fastwam}), which conditions solely on the first frame (reference image) $\mI_\mathrm{ref}$. We incorporate FGP into both regimes by co-denoising a task-level final goal, enabling action tokens to attend directly to it. 
\textbf{(b)} \textbf{Success rates across training-data fractions ($\{0.3,0.5,1.0 \}$)}:   FGP improves performance in most evaluated settings, with larger gains under reduced training-data regimes.}
  \label{fig:method_pilot}
\end{figure*}

\paragraph{\textbf{Backbone and Model Variants}} This section documents the full configuration behind Figure~\ref{fig:pilot_plot}.
All variants are trained and evaluated with the FastWAM codebase, and
all four configurations share an identical WAM setting from~\citep{yuan2026fastwam}:  a
MoT that jointly models future video and robot
actions on a common flow-matching diffusion backbone built on
Wan2.2-TI2V-5B, with a Wan2.1-T2V-1.3B text tokenizer. Each training window contains $33$ observation frames,
encoded by the Wan VAE into a latent video sequence, together with an action
chunk of $32$ steps (an action-to-video frequency ratio of $4$, i.e.\ $9$ video
latent frames per window). Both experts are trained with flow matching (logit-normal timestep shift $5.0$,
$1000$ train timesteps) and share the text conditioning (T5 embeddings,
precomputed and cached per task). Language is the only task conditioning; no
end-effector or goal image is provided as input unless stated otherwise.
The variants below are identical in weights, data, and optimization and
differ only in \emph{how the action expert is coupled to the video}
(its cross-attention structure and denoising schedule) and in whether an
auxiliary goal-prediction target is added. For two baselines,  we adopt $\mathcal{L}_\mathrm{video}$ and $\mathcal{L}_\mathrm{act}$ together during the training, and for FDP, we further add $\mathcal{L}_\mathrm{sub}$ with setting $k=1$ and the sub-goal is treated as the final frame of each task. For inference, here are the differences:

\begin{itemize}[label=$\bullet$, leftmargin=1.5em, itemsep=2pt, topsep=2pt]
  \item \textbf{Zero-imagination Paradigm.} The model denoises video and actions 
        jointly in a single pass. Action tokens cross-attend directly to 
        the initial observation (the first-frame video latents) without relying 
        on imagined future video, generating actions and video in lockstep.

  \item \textbf{Full-imagination Paradigm.} The model decouples video and action 
        generation into a two-stage pipeline. At inference, Stage~1 first 
        generates the future video trajectory from noise, and Stage~2 infers 
        the action chunk conditioned on this generated video via inverse dynamics.

\item \textbf{FGP.} Both paradigm can be augmented with 
        a dedicated final-goal prediction. At inference, the goal and actions are 
        jointly denoised from noise, allowing action tokens to co-attend to 
        the evolving goal representation with first frame, explicitly guiding the policy generation. 
\end{itemize}

\paragraph{\textbf{Benchmarks and Observation Spaces.}} We evaluate our method across three simulation benchmarks spanning single-arm, bimanual, and long-horizon kitchen manipulation: LIBERO, RoboTwin, and RoboCasa~\citep{nasiriany2024robocasa}. Table~\ref{tab:pilot-bench} summarizes their respective observation and action spaces. For our pilot validation, we evaluate on a representative subset of five RoboCasa tasks (\emph{CloseFridge}, \emph{OpenCabinet}, \emph{OpenDrawer}, \emph{TurnOnMicrowave}, and \emph{TurnOnSinkFaucet}). All benchmarks share a unified configuration with $33$ observation frames, an action-to-video ratio of $4$, and a maximum text context length of $128$, differing only in camera count, video resolution, and action/state dimensionality as listed in Table~\ref{tab:pilot-bench}, where we mostly follow~\citet{yuan2026fastwam}.

\begin{table}[ht]
\centering
\small
\setlength{\tabcolsep}{4pt}
\caption{\textbf{Benchmark observation and action spaces.} All use a $33$-frame
window and language
conditioning. ``Concat'' = per-camera frames resized then tiled into a single
video canvas.}
\label{tab:pilot-bench}
\begin{tabular}{lccccc}
\toprule
Benchmark & Tasks & Cameras & Video (Concat) & Action & State \\
\midrule
LIBERO (4 suites) & $4\times10$ & $2$ (agent, wrist)              & $224\times448$ (horiz.) & $7$-D  & $8$-D  \\
RoboTwin 2.0      & $50$        & $3$ (head, L/R wrist)           & $384\times320$          & $14$-D & $14$-D \\
RoboCasa (5-task) & $5$         & $3$ (L/R agent, eye-in-hand)    & $384\times320$          & $12$-D & $16$-D \\
\bottomrule
\end{tabular}
\end{table}

\paragraph{\textbf{Training Configuration.}} The data-ratio sweep sub-samples the \emph{training} set. Given a ratio $R \in \{0.3, 0.5, 1.0\}$, we deterministically shuffle the training episodes with a fixed seed and retain the first $R$ fraction. The resulting subsets are therefore nested ($\mathcal{D}_{0.3} \subset \mathcal{D}_{0.5} \subset \mathcal{D}_{1.0}$), so larger data budgets are strict supersets of smaller ones with no samples dropped. All runs share the optimization hyperparameters. LIBERO-based models are trained on 8 GPUs, while RoboTwin and RoboCasa use up to 64 GPUs. With fixed epochs, total optimizer steps scale linearly with the data ratio.

\paragraph{\textbf{Unifying Distal Planning and Control via Native Goal Prediction.}}
Existing WAMs differ fundamentally in \emph{how much generated future context conditions the action expert}, denoted by $\mathcal{C}(\tX)$. This spectrum spans from \emph{full-imagination} models that condition on full-horizon latent clips ($\mathcal{C}(\tX) = \tX_{1:f}$) to leverage dynamic priors~\citep{25pi0.5,ye2026dreamzero}, down to \emph{zero-imagination} baselines that bypass visual generation entirely during inference ($\mathcal{C}(\tX) = \emptyset$)~\citep{yuan2026fastwam}. 
However, regardless of the choice of $\mathcal{C}(\tX)$, control is ultimately anchored on a compact, short-horizon rollout (video clip $\tV$) starting from the current observation. While such short-sighted conditioning captures local dynamics effectively, it provides minimal long horizon guidance toward the ultimate destination, constraining sample efficiency and generalization under data scarcity. 
While hierarchical VLAs attempt to mitigate this myopia by training dedicated goal experts~\citep{intelligence2026pi07steerablegeneralistrobotic, shou2026halo, long2026scaling}, as a generative model over future latents, we show that WAM natively handles long horizon planning without extra overhead, enabling the video expert to synthesize distal targets that guide the action expert with long-horizon context.

\begin{table}[h]
\centering
\small
\setlength{\tabcolsep}{10pt}
\caption{\textbf{Training hyperparameters for the pilot experiments.} }
\label{tab:pilot-hparams}
\begin{tabular}{cc}
\toprule
Hyperparameter & Value \\
\midrule
Optimizer                & AdamW ($\beta_1{=}0.9$, $\beta_2{=}0.95$) \\
Peak learning rate       & $1\times10^{-4}$ \\
LR schedule              & cosine \\
Epochs                   & $10$-LIBERO, $5$-RoboTwin, $20$-RoboCasa \\
Global batch size        & $128$-LIBERO, $1024$-(RoboTwin, RoboCasa) \\
Gradient accumulation    & $1$ \\
Weight decay             & $1\times10^{-2}$ \\
Max gradient norm        & $1.0$ \\
Flow-matching timesteps  & $1000$ (logit-normal shift $5.0$) \\
Loss weights             & $\lambda_{\text{act}}{=}1.0$, $\lambda_{\text{sub}}{=}1.0$ (FGP) \\
\bottomrule
\end{tabular}
\end{table}

\begin{table*}[t]
  \centering
  \caption{\textbf{Pilot evaluation of FGP.}
   Average SR (\%) under ($30\%$, $50\%$, $100\%$) training-data regimes.}
  \label{tab:fgp_pilot}

  \resizebox{\textwidth}{!}{%
  \begin{tabular}{lccc}
  \toprule
  \multirow{2}{*}{Method}
  & LIBERO
  & RoboTwin
  & RoboCasa (5 Tasks) \\
  \cmidrule(lr){2-2}
  \cmidrule(lr){3-3}
  \cmidrule(lr){4-4}
  & $30\%~/~50\%~/~100\%$
  & $30\%~/~50\%~/~100\%$
  & $30\%~/~50\%~/~100\%$ \\
  \midrule

  Zero-Imagination
  & 82.35 / 90.95 / \textbf{97.40}
  & 79.82 / 88.64 / \textbf{91.71}
  & 50.00 / 60.40 / 75.60 \\

  Zero-Imagination + FGP
  & \textbf{88.72} / \textbf{94.05} / 97.00
  & \textbf{82.57} / \textbf{89.21} / 91.25
  & \textbf{57.20} / \textbf{62.43} / \textbf{76.00} \\

  \midrule

  Full-Imagination
  & 93.55 / 95.75 / 98.25
  & 84.18 / 88.44 / 91.41
  & 68.40 / 70.00 / 82.40 \\

  Full-Imagination + FGP
  & \textbf{94.50} / \textbf{97.85} / \textbf{98.65}
  & \textbf{87.11} / \textbf{89.10} / \textbf{91.54}
  & \textbf{74.00} / \textbf{76.40} / \textbf{87.20} \\

  \bottomrule
  \end{tabular}%
  }
  \end{table*}

\section{Video-only Pretrained Expert} \label{sec_appendix_pre}

\noindent{\textbf{Data Recipe.}} The goal of this stage is to endow the policy with a strong, embodiment-aligned visual dynamics prior before any action learning, so that the subsequent action head can be grounded in an already-capable video predictor rather than learned from scratch. To this end, ProWAM is obtained through a \emph{video-only} pretraining stage that adapts a large-scale video prior to the target robot embodiment before any action learning. Similar to training strategy in~\citet{beingbeyond2025beingh0, wu2026oscar}, we build on the Wan2.2-TI2V-5B video diffusion transformer, initialized from a checkpoint pretrained on large-scale robot video data, a mixture of DROID~\citep{ye2026dreamzero}, AgiBott-Beta ~\citep{bu2025agibot} filtered from~\citep{wu2026oscar}, and 57 tasks from OpenX-Embodiment~\citep{o2024open} with a ratio of  $60\%/30\%/10\%$. This endows general-purpose Wan with a task-level sub-goal prior for robotic manipulation. To monitor generalization we hold out a fixed, seed-deterministic set of 400 episodes (200 DROID + 200 AgiBot), excluded from training, and evaluate the forward-only flow-matching loss on them every 500 steps.

\noindent{\textbf{Resolution and Multi-View Composition.}} Each training clip contains $25$ frames rendered at $512\times512$. Because the source datasets differ in the number of available camera views, we render every episode into a \emph{single composite frame} whose layout is selected automatically from the dataset's native view count (depth streams excluded, and capped at four views): a single view fills the whole frame; two views are placed left$\mid$right (each $\tfrac{W}{2}\times H$); three views use a top-1/bottom-2 layout (view~0 spanning the top half at $W\times\tfrac{H}{2}$, the other two tiled below, each $\tfrac{W}{2}\times\tfrac{H}{2}$); and four views form a $2\times2$ grid (each cell $\tfrac{W}{2}\times\tfrac{H}{2}$). Each view is letterboxed into its cell with aspect ratio preserved (rather than stretched), avoiding spatial distortion. This yields a fixed-size observation interface that handles single- and multi-view episodes uniformly, and lets a single model learn across heterogeneous camera rigs without per-dataset architectural changes.

\begin{figure*}[h]
    \centering
    \includegraphics[width=0.75\linewidth]{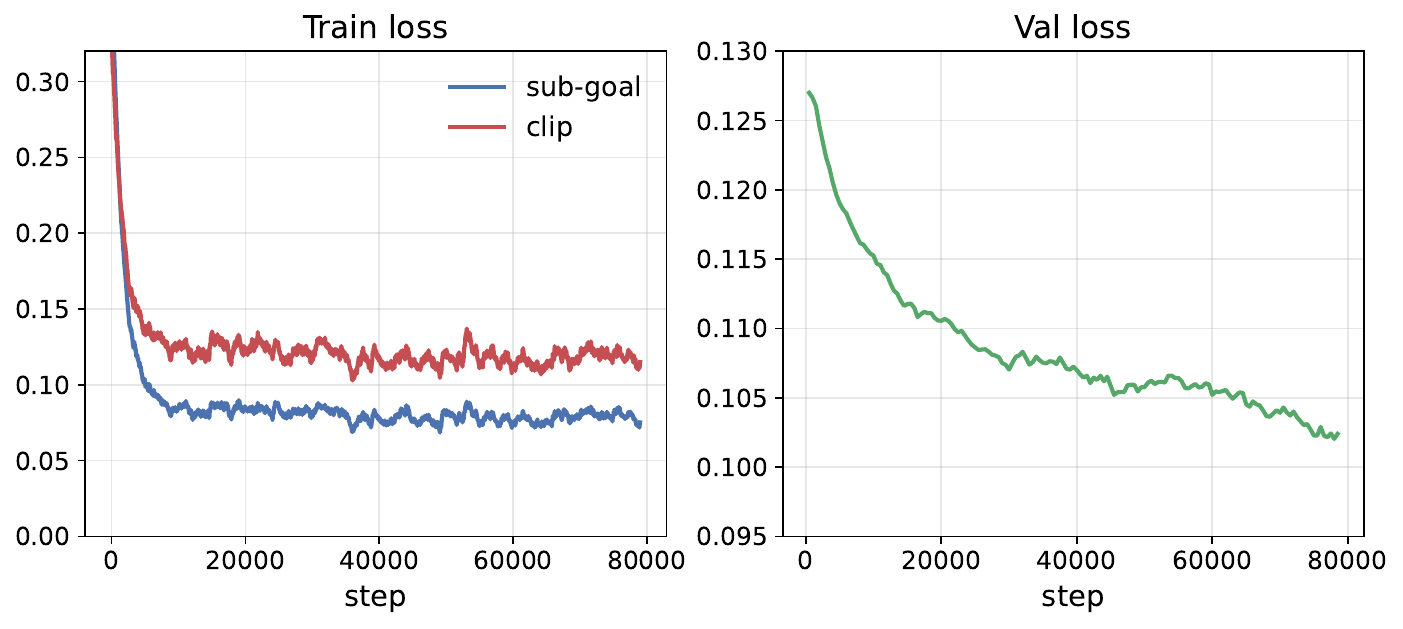}
    \caption{\textbf{Pre-training curves for the sub-goal video expert.} Both training losses (sub-goal and clip) and the validation (Val) loss decrease overall throughout pre-training.}
    \label{fig:pretrain}
\end{figure*}

\paragraph{\textbf{Video-only sub-goal objective.}} At this stage no action head is attached, so the model is trained purely as a flow-matching video predictor. The key ingredient is \emph{progressive sub-goal conditioning}: for each clip we sample up to $K{=}8$ future sub-goal frames from the same trajectory at relative progress positions (via coverage sampling, which spreads sub-goals across the horizon rather than clustering them), and condition the predictor on them jointly with the current observation. Sub-goals are injected with a zeroed rotary position and referenced relative to the first frame, and the clip loss is computed bidirectionally. The sub-goal and clip losses are weighted equally, both set to $1$. Training the video model to forecast frames toward sampled sub-goals is precisely what instills the task-level sub-goal prior that ProWAM later reuses for action prediction. We set $\lambda_{\mathrm{video}} = \lambda_{\mathrm{sub}} = 1$.

\begin{figure*}[h]
    \centering
    \includegraphics[width=1.0\textwidth]{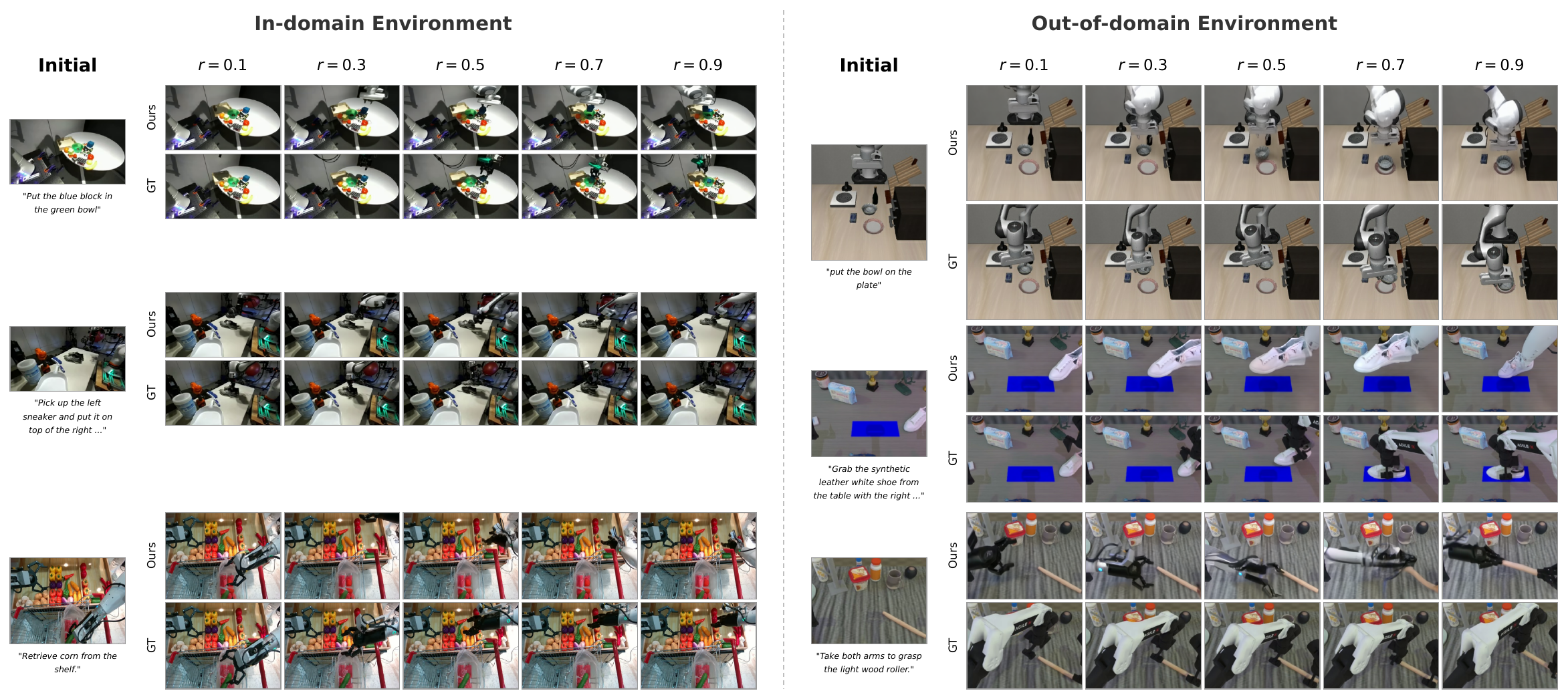}
    \caption{\textbf{Visualized sub-goal prediction of our pretrained video expert}. Each example shows the initial frame (Initial) and predicted sub-goals at five progress values $r\in\{0.1,0.3,0.5,0.7,0.9\}$, with our prediction (Ours, top) above the ground truth (GT, bottom). Left: in-domain training embodiments (DROID, AgiBot). Right: out-of-domain environments unseen during pre-training (LIBERO, RoboTwin).   The selected examples qualitatively illustrate transfer to environments not
  used during video pretraining.}
    \label{fig:pretrain_subgoal}
\end{figure*}

\paragraph{\textbf{Optimization.}} We train only the diffusion transformer (the T5 text encoder and the VAE are kept frozen), on 128 NVIDIA H100 GPUs using DeepSpeed ZeRO-2 sharding with bf16 mixed precision. We use AdamW ($\beta_1{=}0.9$, $\beta_2{=}0.999$, weight decay $10^{-2}$) with a constant learning rate of $1\times10^{-5}$ and a global batch size of $256$ for $8$ epochs ($\sim$79k steps, i.e.\ roughly $2\times10^{7}$ video clips seen). 

\paragraph{\textbf{Loss Analysis.}}   Figure~\ref{fig:pretrain} summarizes the pretraining dynamics. Both the clip and sub-goal training losses decrease rapidly during the first ${\sim}10{,}000$ steps before approaching stable plateaus. Over the reported 79k-step schedule, the held-out flow-matching loss decreases overall from $0.127$ to $0.102$, with no sustained increase. We therefore use the final checkpoint for downstream fine-tuning.

\paragraph{\textbf{Sub-goal Generation.}}  As shown in Figure~\ref{fig:pretrain_subgoal}, the pretrained video expert predicts sub-goals that advance coherently with the progress value $r$. As $r$
  increases from $0.1$ to $0.9$, the generated frames depict the manipulation
  progressing toward task completion, such as approaching, grasping, and
  transporting the target object, and qualitatively align with the corresponding
  ground-truth frames on the training embodiments. Similar behavior is observed
  on LIBERO and RoboTwin, which are not included during pretraining: the predicted sub-goals remain instruction-aligned and capture major object interactions and arm motions. These qualitative results suggest that progress-conditioned sub-goal prediction transfers to unseen embodiments and scenes, providing useful visual context for the downstream action expert.

\section{Simulation Experiments}\label{sec_app_downstream}

\paragraph{\textbf{Training.}} During downstream adaptation, all the simulation benchmarks adopt a two-stage training
paradigm on top of the pretrained sub-goal video expert. \textbf{Stage 1
(Cold-start)} continues training only the video expert or base Wan 
on the task-specific downstream video data (no action head), keeping the progressive-sub-goal
conditioning, so the model adapts to the target visual domain while preserving its learned future-prediction prior. \textbf{Stage 2 (Joint SFT)} attaches the action DiT and co-trains video and action with a mixed-attention coupling, where the action tokens attend to the current frame and the predicted sub-goal(s). We set $\lambda_{\mathrm{video}} = \lambda_{\mathrm{sub}} = \lambda_{\mathrm{act}} = 1$. The simulation benchmarks share this recipe and differ only in data and clip length, all hyper-parameters are listed in Table~\ref{tab:downstream_hparams}- \ref{tab:downstream_hparams_robocasa}:

\begin{itemize}[label=$\bullet$, leftmargin=1.5em, itemsep=2pt, topsep=2pt]
\item \textbf{LIBERO \& LIBERO-Plus.} Standard LIBERO~\citep{LIBERO23} consists of four single-arm tabletop manipulation suites (Spatial, Object, Goal, and Long), each featuring 10 tasks and 500 expert demonstrations. We train our policy across all four standard suites and report the \emph{average success rate} (ASR) over 50 evaluation trials per task, yielding a total of 2000 trials across 40 tasks. LIBERO-Plus~\citep{fei25libero-plus} introduces a more challenging setting built upon the original LIBERO tasks, incorporating diverse perturbations (camera, robot, language, lighting, background, sensor noise, and layout). We use only the original LIBERO demonstrations for training and do not incorporate any LIBERO-Plus data, which is used exclusively for zero-shot generalization evaluation. We evaluate under the LIBERO-Plus protocol with ASR. We use all four LIBERO suites, with the original 50 demonstrations per task. Observations are the agent-view and wrist cameras composited side-by-side at $224\times448$. Stage~1 performs video-only adaptation from the pretrained expert for 9 epochs, the resulting checkpoint initializes the video expert for Stage 2, which co-trains the action head for another 10 epochs. Unless otherwise noted, our best model uses a single sub-goal ($k{=}1$) at progress $r{=}0.7$.

\item {\textbf{RoboTwin.}} RoboTwin~\citep{chen2025robotwin} is a large-scale simulated benchmark for bimanual robot manipulation, which covers 50 tasks and requires policies to coordinate under diverse object layouts and scene conditions. Specifically, it provides 2,500 trajectories collected in clean scenes and 25,000 trajectories collected with heavy scene randomization. In contrast to prior work that trains on both~\citep{yuan2026fastwam,zhang2026imagewam,li2026lingbotva}, we train our policy exclusively on the 2,500 clean-scene demonstrations (Clean) and treat the randomized scenes (Random) as an out-of-distribution test to scrutinize the model's cross-domain generalization, reporting ASR over 100 trials per task. We use all 50 RoboTwin tasks under the multi-task setting, training on the 50 clean demonstrations per task. The randomized split evaluates control generalization without randomized-scene action supervision. Following~\citep{chen2026mv,hu2026bagelvla}, the video expert is trained with action-free videos from both clean and randomized scenes, whereas vision-action fine-tuning uses only clean demonstrations. The randomized split therefore evaluates whether visual knowledge learned without action labels can support action generation under substantial scene variation. Observations are the head and two wrist cameras composited into a single frame at $384\times320$ (head on the top two-thirds, left/right wrist on the bottom). We follow the same two-stage recipe as LIBERO, with three differences: (i) a three-camera RoboTwin composite instead of two, (ii) a small cold-start (RoboTwin is roughly $24\times$ larger than LIBERO, so video-only adaptation runs about $1$ epoch), and (iii) the best sub-goal schedule is near-biased with a higher $k$ ($K{=}4$, progress ${0.1,0.2,0.35,0.5}$). Stage 2 co-trains the action head for 5 epochs on the clean split with the same differential learning rate.

\item {\textbf{RoboCasa365.}} RoboCasa365~\citep{nasiriany2024robocasa} is a large-scale benchmark for generalist robot policies, spanning 365 everyday manipulation tasks across 2,500 diverse kitchen environments. Following the official multi-task protocol, policies are trained on the Human300 dataset, which contains human-teleoperated demonstrations for 300 tasks, and evaluated in closed loop on 50 target tasks. The evaluation comprises 18 \emph{Atomic-Seen}, 16 \emph{Composite-Seen}, and 16 \emph{Composite-Unseen} tasks. The first two splits contain tasks observed during pretraining, whereas \emph{Composite-Unseen} evaluates zero-shot generalization to held-out task compositions. We report success rates for each split and the mean success rate over all 50 tasks as the overall score. We follow the same two-stage recipe used for LIBERO~\citep{LIBERO23} and RoboTwin~\citep{chen2025robotwin}. As shown in Table~\ref{tab:downstream_hparams_robocasa}, in Stage~1, we pretrain the video expert on the RoboCasa demonstrations with the progressive sub-goal objective; in Stage~2, we jointly fine-tune the video expert and the action expert. The three simulated views (one top, two bottom) are composited into a single $448 \times 448$ image (top view across the upper region, wrist/side views below). Each clip spans 49 raw frames sampled at stride 3 (17 latent frames). Actions ($d_a = 12$) and proprioception ($d_p = 16$) are normalized using dataset statistics, and the action expert predicts a horizon of $H = 32$. Sub-goals use a progressive-$K$ schedule with $K_{\max} = 6$ and progress levels $r \in \{0.3, 0.4, 0.5, 0.65, 0.8, 0.9\}$. We train at a global batch size of 1024 with a cosine schedule, learning rate $1 \times 10^{-4}$ for the action expert and $5 \times 10^{-5}$ for the video expert. We run the official RoboCasa closed-loop evaluator on three splits:
\emph{atomic-seen} ($18$ tasks), \emph{composite-seen} ($16$ tasks), and
\emph{composite-unseen} ($16$ tasks), with $25$ independent rollouts per task
($450$/$400$/$400$ trials respectively). At inference, the video expert is evaluated once per replanning step to obtain and cache the sub-goal key-value features. The action expert then reuses this cache over $10$ action-denoising steps. The cache is refreshed when a new observation triggers the next replanning step, and executes the \emph{full} predicted horizon before
re-planning ($H{=}32$, replan every $32$ steps). We report the
per-task success rate averaged over trials for each split. 
\end{itemize}
\begin{table}[t]
\centering
\small
\caption{\textbf{Downstream two-stage training hyperparameters (LIBERO).}}
\label{tab:downstream_hparams}
\begin{tabular}{lcc}
\toprule
 & \textbf{Stage 1 (Cold Start)} & \textbf{Stage 2 (Joint SFT)} \\
\midrule
Trainable            & video expert    & video expert + action head \\
Action horizon       & --                   & 16 \\
Subgoals ($k$)       & Random $k\sim\mathcal{U}\{1,\dots,8\}$ & $k=1$ ($\{0.7\}$)\\
Learning rate        & $1\times10^{-4}$  & $1\times10^{-4}$ (action), $5\times10^{-5}$ (video) \\
Scheduler            & constant             & cosine, 5\% warm-up \\
Global batch         & 64                   & 128 \\
Optimizer & \multicolumn{2}{c}{AdamW ($\beta_2 = 0.95$, weight decay $10^{-2}$)} \\
Grad clip            & \multicolumn{2}{c}{1.0} \\
Epochs       & \multicolumn{2}{c}{10} \\
\bottomrule
\end{tabular}
\end{table}

\begin{table}[t]
\centering
\small
\caption{\textbf{Downstream two-stage training hyperparameters (RoboTwin).}}
\label{tab:downstream_hparams_robotwin}
\begin{tabular}{lcc}
\toprule
& \textbf{Stage 1 (Cold Start)} & \textbf{Stage 2 (Joint SFT)} \\
\midrule
Trainable & video expert & video expert + action head \\
Action horizon & -- & 16 \\
Subgoals ($k$) & Random $k \sim \mathcal{U}\{1,\dots,8\}$ & $k = 4$ ($\{0.1, 0.2, 0.35, 0.5\}$) \\
Learning rate & $5\times10^{-5}$  & $1\times10^{-4}$ (action), $5\times10^{-5}$ (video) \\
Scheduler & constant & cosine, 5\% warm-up \\
Epochs       & 1 & 5 \\
Optimizer & \multicolumn{2}{c}{AdamW ($\beta_2 = 0.95$, wd $10^{-2}$)} \\
Grad clip & \multicolumn{2}{c}{1.0} \\
Global batch & \multicolumn{2}{c}{1024} \\
\bottomrule
\end{tabular}
\end{table}

\begin{table}[t]
\centering
\small
\caption{\textbf{Downstream two-stage training hyperparameters (RoboCasa365).}}
\label{tab:downstream_hparams_robocasa}
\begin{tabular}{lcc}
\toprule
& \textbf{Stage 1 (Cold Start)} & \textbf{Stage 2 (Joint SFT)} \\
\midrule
Trainable & video expert & video expert + action head \\
Action horizon & -- & 32 \\
Subgoals ($k$) & Random $K \sim \mathcal{U}\{1,\dots,8\}$ & $k = 6$ ($\{0.3, 0.4, 0.5, 0.65, 0.8, 0.9\}$) \\
Learning rate & $1\times10^{-5}$ & $1\times10^{-4}$ (action), $5\times10^{-5}$ (video) \\
Scheduler & constant & cosine, no warm-up \\
Epochs & $0.8$ (cold start) & 1.5 \\
Global batch & 64 & 1024 \\
Resolution & \multicolumn{2}{c}{$448\times448$, 3-view composite (top-1 / bottom-2)} \\
Frames & \multicolumn{2}{c}{49 raw @ stride 3 ($\rightarrow$ 17 latent frames)} \\
Optimizer & \multicolumn{2}{c}{AdamW ($\beta_1 = 0.9$, $\beta_2 = 0.999$, wd $10^{-2}$)} \\
Grad clip & \multicolumn{2}{c}{1.0} \\
\bottomrule
\end{tabular}
\end{table}

\begin{table}[h]
  \centering
\caption{\textbf{Closed-loop SR (\%) on RoboCasa365}. ProWAM ranks \textbf{4th overall} on the benchmark (updated at 10/01/2026), outperforming most existing policies especially on unseen-composite tasks.
}
  \label{tab:robocasa_main}
  \resizebox{\linewidth}{!}{
  \begin{tabular}{clcccc}
    \toprule
    Ranking & Method & Overall & Atomic-Seen & Composite-Seen & Composite-Unseen \\
    \midrule
    1 & Paimon-0           & 58.1 & 79.3 & 55.8 & 36.7 \\
    2 & Xiaomi-Robotics-1~\citep{team2026xiaomi}    & 57.4 & 80.2 & 57.1 & 32.1 \\
    3 & Phasor-m7    & 56.8 & 78.7 & 57.4 & 31.5 \\
    \rowcolor{gray!12} 
    4 & \textbf{ProWAM (Ours)} & \textbf{48.1} & \textbf{76.0} & \textbf{46.5} & \textbf{18.2}
    \\
    5 & ABot-M0.6~\citep{chen2026abot}            & 46.6 & 79.4 & 48.3 & 7.9 \\
    6 & ABot-M0.5~\citep{chen2026abot}            & 40.3 & 75.6 & 37.7 & 3.3 \\
        7 & PRTS~\citep{zhang2026prtsprimitivereasoningtasking}                & 39.6 & 66.3 &	30.3&	18.8 \\
    8 & RLDX-1~\citep{kim2026rldx}                & 36.0 & 67.6 & 27.9 & 8.5 \\
    
    9 & WorldDreamer~\citep{worldagents2026worlddreamer}         & 35.3 & 66.3 & 26.7 & 9.0 \\
    10 & GR00T N1.5~\citep{bjorck2025gr00t}           & 23.9 & 50.7 & 14.8 & 2.7 \\
    10 & GR00T N1.6~\citep{bjorck2025gr00t}            & 21.9 & 51.1 & 9.4  & 1.7 \\
    11 & GigaWorld-Policy 0.1~\citep{ye2026gigaworldpolicy} & 20.7 & 44.4 & 11.8 & 2.9 \\
    12 & $\pi_{0.5}$~\citep{25pi0.5}          & 16.9 & 39.6 & 7.1  & 1.2 \\
    13 & $\pi_0$~\citep{24pi0}           & 14.8 & 34.6 & 6.1  & 1.1 \\
    \bottomrule
  \end{tabular}
  }
\end{table}

\begin{table}[h]
\centering
\small
 \caption{\textbf{Impact of KV-cached inference and explicit sub-goal
  conditioning on LIBERO.}}
\label{tab:inference_ablation}
\setlength{\tabcolsep}{7pt}
\begin{tabular}{l c cc c}
\toprule
\textbf{Method} & \textbf{Explicit Sub-goal} & \textbf{ID} & \textbf{OOD} & \textbf{TFLOPs} \\
\midrule 
ProWAM (Joint-10)    & \checkmark & \textbf{99.0} & 85.8 & 113.4 \\
ProWAM (KV-3)     & \checkmark & 98.8 & 85.6 & 20.6  \\ %
ProWAM (KV-1)     & \checkmark & 98.8 & \textbf{85.8} & 15.1  \\
\midrule
ProWAM-Light       & $\times$   & 98.6 & 80.3 & \textbf{13.8} \\
\bottomrule
\end{tabular}
\end{table}

\paragraph{\textbf{Efficient Inference.}} 

\begin{figure*}[t]
    \centering
    \includegraphics[width=1.0\textwidth]{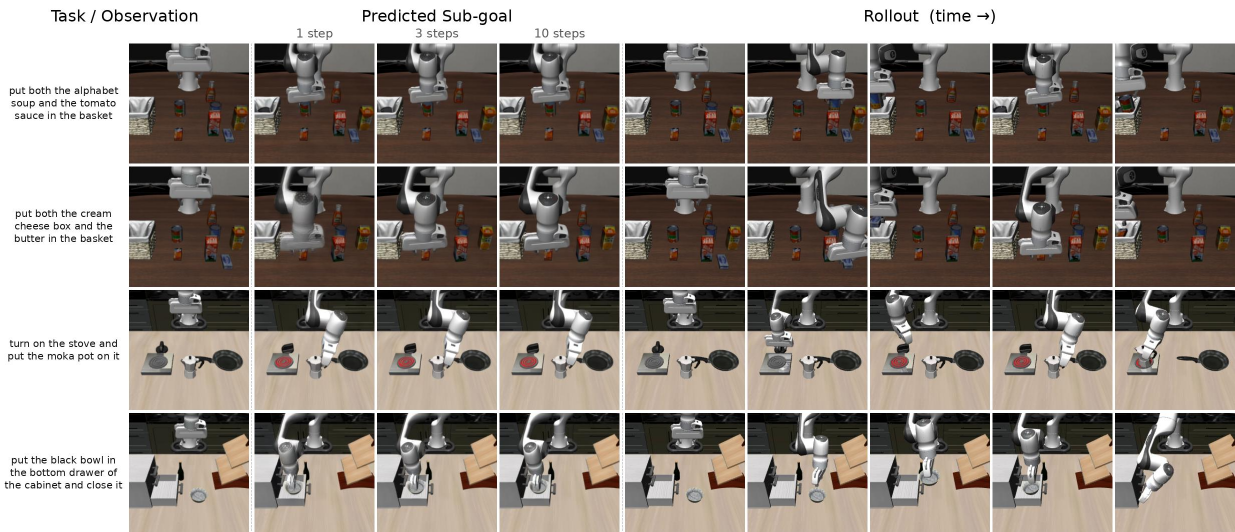}
    \caption{\textbf{Qualitative results on LIBERO}. Left: the first frame observation + instruction. Middle: imagined sub-goal after $1/3/10$ denoising steps. Right: executed rollout.}
    \label{fig:pretrain_subgoal_libero}
\end{figure*}

  A naive world-action policy denoises the imagined sub-goal and action jointly
  for the full $T{=}10$ diffusion steps, invoking the video expert at every step.
  We find that this repeated video-expert computation is unnecessary in our
  experiments. We first inspect the generated sub-goals after different numbers of
  denoising steps (Fig.~\ref{fig:pretrain_subgoal_libero}). Additional denoising
  improves visual detail, consistent with observations in
  FastWAM~\citep{yuan2026fastwam} and
  ImageWAM~\citep{zhang2026imagewam} that photorealistic predictions are not
  required to condition the policy.
  
  We therefore decouple the two branches at inference. At each replanning step,
  the video expert performs a single forward pass to predict the sub-goals and
  cache their key--value features. The action head then reuses this fixed cache
  throughout $T$ action-denoising steps, with no additional video-expert forwards.
  This reduces video-expert evaluations from $T$ to $1$ ($\approx10\times$
  fewer evaluations), yielding an inference cost comparable to FastWAM and
  ImageWAM. As shown in Table~\ref{tab:inference_ablation}, KV-1 and Joint-10
  achieve the same observed OOD success rate ($85.8\%$), while KV-1 reduces
  computation by $86.7\%$. This suggests that sub-goal features from a single
  video-expert evaluation provide sufficient conditioning for action denoising in
  this setting.

\paragraph{\textbf{Beyond Explicit Sub-goal Conditioning}} \label{sec_app_prowamlight}
  To test whether the action head can rely on the visual knowledge encoded by the
  pretrained video backbone without explicit sub-goal conditioning, we introduce
  \textbf{ProWAM-Light}, inspired by FastWAM's fast-inference setting. ProWAM-Light
  masks sub-goal tokens during both training and inference, such that the action
  head conditions only on the current observation. At inference, a single
  video-expert forward pass constructs a current-observation key--value cache that
  is reused throughout action denoising. As shown in
  Table~\ref{tab:inference_ablation}, ProWAM-Light maintains high in-domain
  performance, but its OOD success decreases from $85.8\%$ to $80.3\%$ despite
  only a modest compute reduction ($15.1$ to $13.8$ TFLOPs), demonstrating the
  benefit of explicit sub-goal conditioning for generalization.

\section{Real Robotic Experiment}~\label{sec_app_Real Robotic Experiment}

\begin{figure}[t]
  \centering
\begin{minipage}[c]{0.25\linewidth}
  \centering
  \includegraphics[height=3.87cm, width=\linewidth, keepaspectratio]{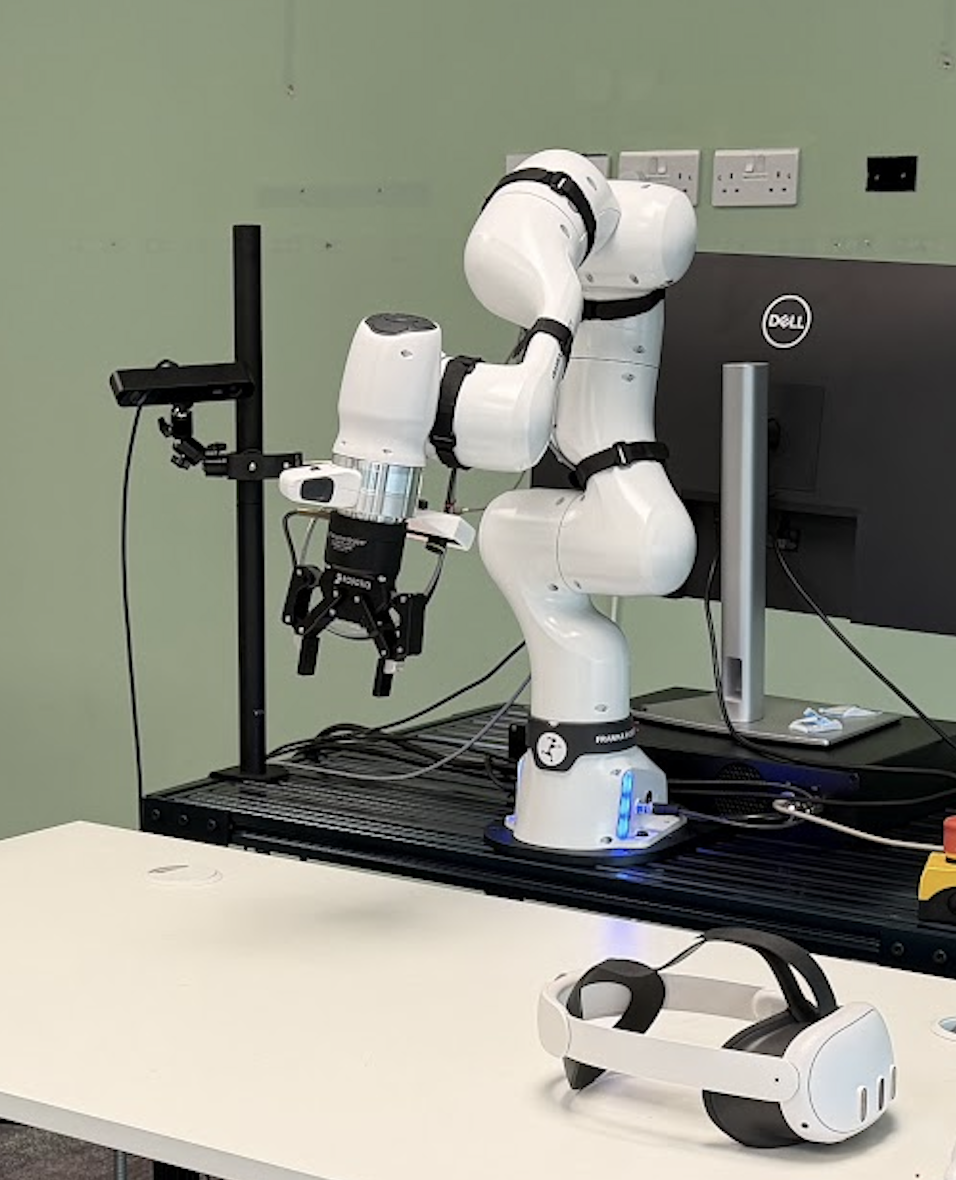}
  \par\vspace{2pt}
  {\footnotesize \textbf{(a) Physical Setup.}}
\end{minipage}
  \begin{minipage}[c]{0.69\linewidth}
    \centering
    \includegraphics[height=3.93cm, width=\linewidth]{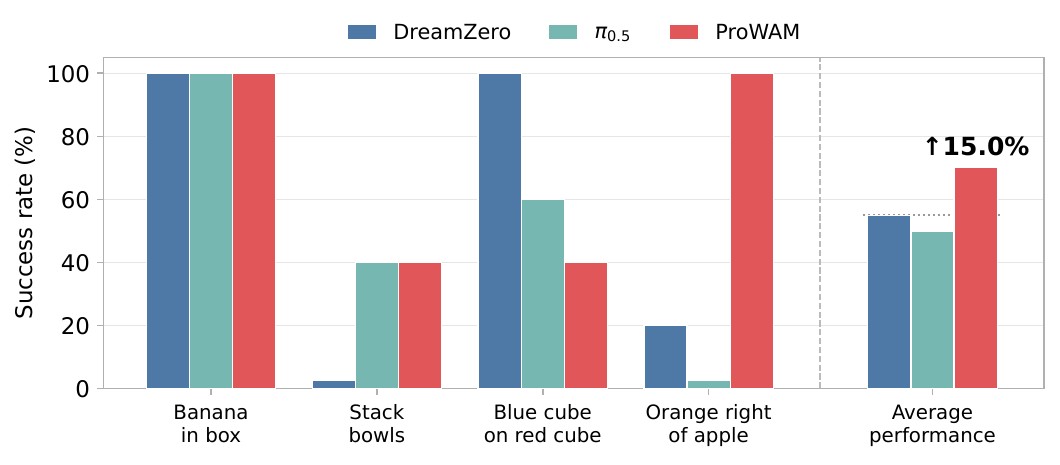}
    \par\vspace{-1pt}
    {\footnotesize (b) \textbf{Real-world Experiment Performance.}}
  \end{minipage}
  \caption{\textbf{Zero-shot real-world evaluation on Franka FR3.} 
   \textbf{(a) The physical evaluation environment setup:} It features a clean tabletop workspace and 3 recorded camera views.
   \textbf{(b) Success rate across $4$ physical tasks:}  ProWAM achieves the highest average success rate of \textbf{70.0\%}, outperforming the strongest baseline by an absolute margin of \textbf{15.0}.}
  \label{fig:real_results}
\end{figure}

\paragraph{\textbf{Implementation Details.}}
To test whether ProWAM   transfers from DROID demonstrations to unseen real-world tasks and scene configurations, we
train on DROID~\citep{khazatsky2024droid}, so every
number in this section measures cross-domain generalization rather than in-domain fitting, e.g., LIBERO and RoboTwin, where training and evaluation share a domain. We follow the same two-stage recipe
used in those experiments. As shown in Table~\ref{tab:downstream_hparams_droid}, in Stage~1
we pretrain the video expert on DROID with the progressive sub-goal objective, in Stage~2 we
jointly fine-tune the video expert and the action expert under a differential learning rate.
The three DROID views (left wrist, two exterior) are composited into a single $576 \times 640$
image, with the wrist view spanning the upper two-thirds and the two exterior views tiled
below. Each clip spans 33 frames at stride $1$.   Actions and proprioception are represented in absolute joint space and use identity normalization. At each replanning step, the policy predicts an action chunk of $H=24$ absolute joint-position targets, executes the first $H_{\mathrm{ol}}=24$ targets open-loop, and then replans from a new observation. The executed targets are sent directly to the robot's joint-position controller at 15\,Hz. Sub-goals follow a balanced near-range ladder with $K_{\max} = 6$: $K$ is drawn uniformly from $1$ to $6$ and the sub-goal positions for each$K$ are placed on a grid spanning $r \in [0.1, 0.5]$. Other settings are similar to previous ones.

\begin{table}[ht]
\centering
\small
\caption{\textbf{Downstream two-stage training hyperparameters on DROID.}}
\label{tab:downstream_hparams_droid}
\begin{tabular}{lcc}
\toprule
& \textbf{Stage 1 (Cold Start)} & \textbf{Stage 2 (Joint SFT)} \\
\midrule
Trainable & video expert & video expert + action head \\
Action horizon & -- & 24 \\
Sub-goals ($k$) & Random $k \sim \mathcal{U}\{1,\dots,8\}$ & $k$ = 6 $(\{0.1, 0.18, 0.26, 0.34, 0.42, 0.5\}$) \\
Learning rate & $1\times10^{-4}$ & video expert $5\times10^{-5}$ / action head $1\times10^{-4}$ \\
Epochs & 0.5 & 2 \\
Resolution & \multicolumn{2}{c}{$576\times640$, 3-view composite (wrist top / 2 exterior bottom)} \\
Optimizer & \multicolumn{2}{c}{AdamW ($\beta_1 = 0.9$, $\beta_2 = 0.999$, wd $10^{-2}$)} \\
Grad clip & \multicolumn{2}{c}{1.0} \\
Scheduler & \multicolumn{2}{c}{cosine, no warm-up} \\
Global batch & \multicolumn{2}{c}{256} \\

\bottomrule
\end{tabular}
\end{table}

\begin{table}[ht]
\centering
\small
\caption{\textbf{Real-world task suite and success criteria.}}
\label{tab:real_task_spec}
\begin{tabularx}{\linewidth}{@{}l p{0.32\linewidth} X@{}}
\toprule
\textbf{Task} & \textbf{Instruction} & \textbf{Success criterion} \\
\midrule
Banana
in box & \emph{Put the banana in the box.} &
The banana comes to rest inside the box. \\
\addlinespace
Stack
bowls & \emph{Stack the bowls.} &
All three bowls are nested into a single stack, with no bowl knocked over or
displaced from the stack at any point during execution. \\
\addlinespace
Blue cube
on red cube & \emph{Put the blue cube on the red cube.} &
The blue cube rests unsupported on the red cube. A release at any stage after
the grasp that does not land the cube on the red cube counts as a failure. \\
\addlinespace
Orange right
of apple & \emph{Put the orange to the right side of the apple.} &
The orange is released to the right of the apple within a reasonable offset.
Position is scored at the first release location, subsequent corrections do not
count. \\
\bottomrule
\end{tabularx}
\end{table}

\begin{figure}[ht]
  \centering
  \includegraphics[width=1.0\linewidth]{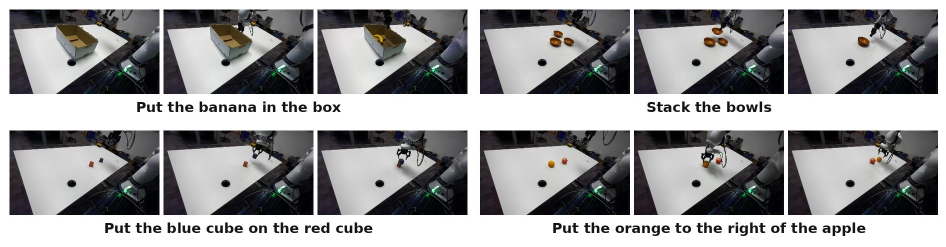}
  \caption{\textbf{Real-world physical evaluation tasks.} Multi-camera observations (2 exterior and 1 wrist-mounted camera, only 1 exterior view is shown here to anonymize individuals) across 4 real-world tasks: \emph{Banana in box}, \emph{Stack bowls}, \emph{Blue cube on red cube}, and \emph{Orange right of apple}.}
  \label{fig:real_tasks_2x2}
\end{figure}

\begin{figure*}[t]
    \centering
    \includegraphics[width=\textwidth]{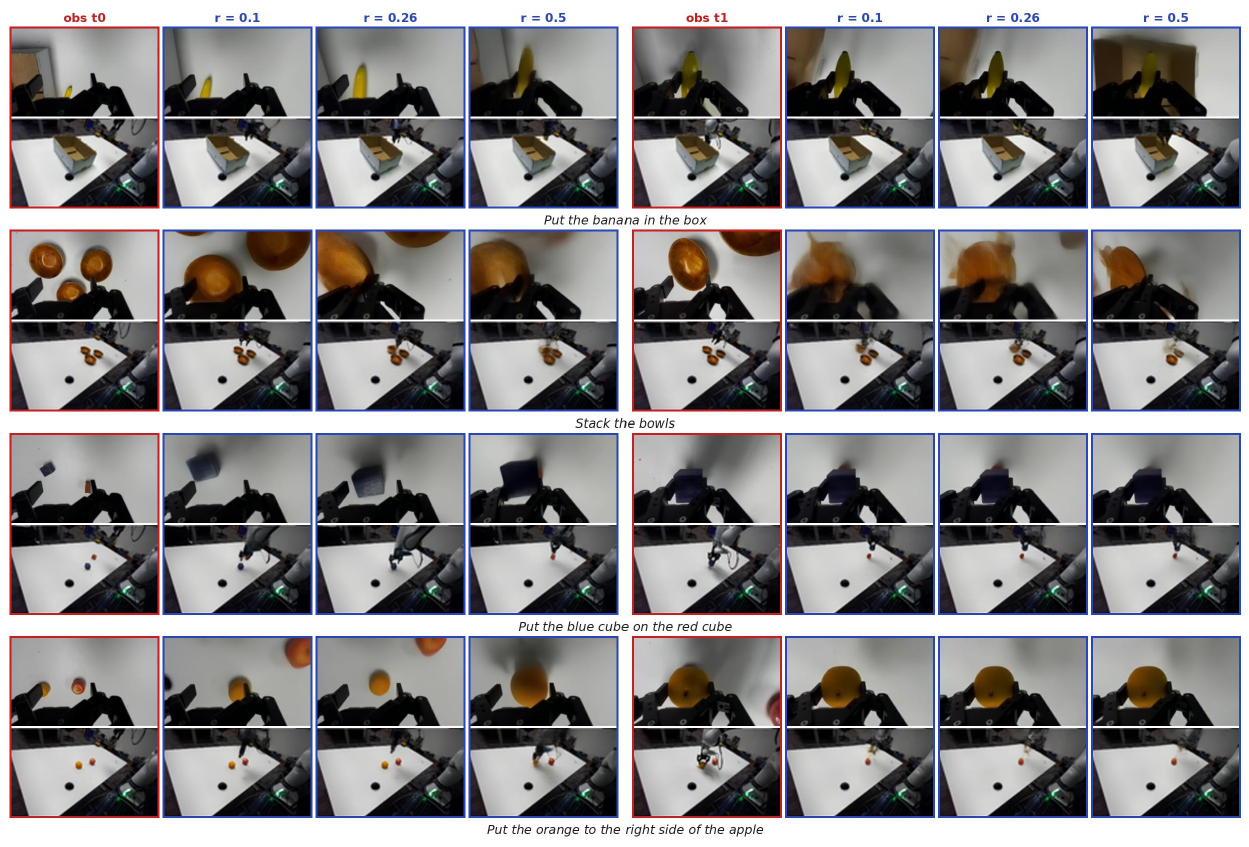}
    \caption{\textbf{Closed-loop visual sub-goal generation in real-world experiments.}
    Given observations (\textcolor{red}{$t_0$}, \textcolor{red}{$t_1$}),
    our model generates progress-conditioned visual sub-goals in a single pass
    at \textcolor{blue}{$r\!\in\!\{0.1, 0.26, 0.5\}$}.
  Although visually coarse, the generated sub-goals illustrate task-relevant
  motion and spatial transitions.}
    \label{fig:real_world_subgoals}
\end{figure*}

\paragraph{\textbf{Real-world Experimental Settings \& Analysis.}}
We deploy the policy on a hardware and software setup that mirrors the original DROID setup~\citep{khazatsky2024droid}. We use a single Franka FR3 arm with a Zed Mini wrist-mounted camera and two Zed 2i exterior cameras. Monocular images used for inference are sourced from the left-rectified sensor from each camera to match the original DROID dataset. We evaluate on a Franka FR3 under randomized initial object configurations, with $5$ trials
per task. The four tasks are designed independently of the RoboLab
suite and span basic pick-and-place, long-horizon multi-object assembly, contact-rich
placement, and spatially grounded placement. All success criteria are fixed prior to evaluation
and applied identically to every method, detailed evaluation criterion is in Table~\ref{tab:real_task_spec}. Figure~\ref{fig:real_world_subgoals} illustrates closed-loop visual sub-goal
generation in our real-world experiments. Given the current observation, our
model produces a temporally ordered visual plan across multiple task-progress
levels, which guides subsequent action generation and enables more effective
policy execution. While the generated sub-goals place less emphasis on
fine-grained textures and photorealistic details, they accurately capture the
task-relevant motion, object interactions, and spatial relationships required
to complete the manipulation tasks.

\section{Simple ABC Experiment}\label{sec_app_abc}

\paragraph{\textbf{Implementation Details.}}   We additionally evaluate ProWAM on ABC, a simulated bimanual manipulation benchmark that models a realistic dual-arm setup with two 7-DoF arms and head- and wrist-mounted cameras. Its open-source closed-loop evaluation protocol (\texttt{bottles-75k}) enables reproducible testing of long-horizon, precision-sensitive manipulation. To this end, we follow a similar training strategy to LIBERO~\citep{LIBERO23} and
RoboTwin~\citep{chen2025robotwin} for training on ABC. For the bottles task, we train on the dataset's native mixture of teleoperated real-robot ($\approx 43\%$) and simulation ($\approx 57\%$) demonstrations provided by ABC. The three cameras (head + two wrists) are composited into a single
$384\times320$ image (head across the top two-thirds, wrists on the bottom
third). Actions and proprioception are z-score
normalized using dataset statistics. We use one sub-goal $k{=}1$ at progress-$r{=}0.7$. 

\paragraph{\textbf{Evaluation protocol.}} We perform the closed-loop evaluation using the official ABC evaluator. Each run covers 20 independent worlds, and each rollout is 120 action chunks of 15 executed steps. We report the strict
AVG (all six bottles placed in the bin) and the mean maximum
number of bottles placed (out of 6). Both our policy and the official policy are evaluated on the same 20 seed-matched worlds, and both face the same training and evaluation renderer gap. Notably, ABC does not publish a simulation success number in either the paper or the code release, we obtained the reference by running their own checkpoint through the same harness. 

\begin{figure}[h]
\centering
\begin{minipage}[c]{0.3\linewidth}
  \centering
  \captionof{table}{Performance on ABC put-bottles-in-bin Task. }
  \label{tab:abc_main}
  \resizebox{\linewidth}{!}{%
  \begin{tabular}{lcc}
  \toprule
  Method & ASR & Mean /6 ($\uparrow$) \\
  \midrule
  ABC official & 0.05 & 4.10 \\
  \textbf{Ours} & \textbf{0.35} & \textbf{4.85} \\
  \bottomrule
  \end{tabular}}
\end{minipage}\hfill
\begin{minipage}[c]{0.65\linewidth}
  \centering
  \includegraphics[width=\linewidth]{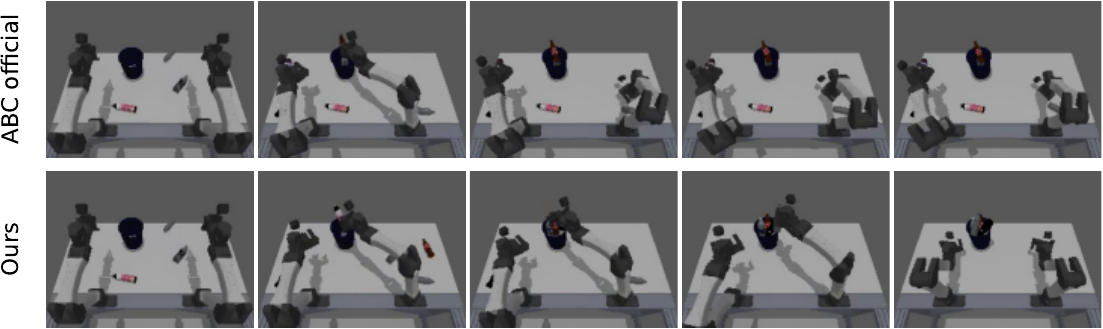}
  \captionof{figure}{\textbf{Visual Comparison between ProWAM and Baseline.}
  Top shows the baseline policy stalls early, leaving bottles unplaced
  (4/6). Bottom shows that our policy clears the entire table
  (6/6).}
  \label{fig:abc_qual}
\end{minipage}
\end{figure}

\paragraph{\textbf{Results.}} Table~\ref{tab:abc_main} reports the main comparison. Our policy succeeds in 7 of 20 evaluated worlds, compared with 1 of 20 for the official ABC policy, and achieves a higher mean bottle count (4.85 versus 4.10). The absolute numbers are modest because the strict 6/6 criterion is
demanding and a training-to-eval renderer gap depresses all policies.
Figure~\ref{fig:abc_qual} contrasts the two policies on the same evaluation
world: our policy places all bottles (bottom row), whereas the official
policy stalls with 2 bottles unplaced (top row). The dominant failure mode across policies lies in precision bottlenecks on the \emph{last} bottle, where the end-effector approaches but fails to secure it.

\section{Limitations and Future Work}
\label{sec:appendix_failure_limitation}
While ProWAM demonstrates strong performance across diverse benchmarks, our study has several limitations that open promising avenues for future work:
\begin{itemize}[label=$\bullet$, leftmargin=1.5em, itemsep=2pt, topsep=2pt]
  \item \textbf{Manually Specified Inference Schedule.}
  Although ProWAM supports variable sub-goal counts and progress values, the
  inference schedule is manually specified rather than inferred from the current
  task state or complexity. Learning to select the number and placement of
  sub-goals adaptively is a promising direction for future work.
    \item \textbf{Model and Pre-training Scale:} We build on a moderately sized video expert and a compact action expert. Due to compute constraints, we do not explore how scaling up the backbone size or pre-training corpora impacts performance, which we leave for future exploration.
    \item \textbf{Task Diversity:}   Although our evaluation includes several multi-stage tasks, the tested horizons
  remain limited compared with extended real-world deployments, and evaluation on more dexterous tasks remains an important direction for future
  work.
\end{itemize}

\end{document}